\documentclass{article}

\PassOptionsToPackage{numbers, compress}{natbib}
    \usepackage[preprint]{neurips_2025}

\usepackage[utf8]{inputenc} %
\usepackage[T1]{fontenc}    %
\usepackage{hyperref}       %
\usepackage{url}            %
\usepackage{booktabs}       %
\usepackage{amsfonts}       %
\usepackage{nicefrac}       %
\usepackage{microtype}      %
\usepackage[table]{xcolor}  %
\usepackage{graphics}
\usepackage{times}              
\usepackage{amsmath}
\usepackage{amssymb}
\usepackage{bm}
\usepackage[noend]{algpseudocode}
\usepackage{color}
\usepackage{subcaption}
\usepackage{rotating}
\usepackage{xstring}
\usepackage{hhline}
\usepackage{siunitx}
\usepackage{tikz}
\usepackage{soul}
\usepackage{tcolorbox}
\usepackage{textcomp}
\usepackage{array,multirow,graphicx}
\usepackage{stmaryrd}
\usepackage{multicol}
\usepackage{hyperref}
\usepackage{xspace}
\usepackage{cleveref}
\usepackage{wrapfig}

\newcommand{\bdmath}{\begin{dmath}}
\newcommand{\edmath}{\end{dmath}}
\newcommand{\beq}{\begin{equation}}
\newcommand{\eeq}{\end{equation}}
\newcommand{\bdm}{\begin{displaymath}}
\newcommand{\edm}{\end{displaymath}}
\newcommand{\bea}{\begin{eqnarray}}
\newcommand{\eea}{\end{eqnarray}}
\newcommand{\beal}{\beq \begin{array}{ll}}
\newcommand{\eeal}{\end{array} \eeq}
\newcommand{\beas}{\begin{eqnarray*}}
\newcommand{\eeas}{\end{eqnarray*}}
\newcommand{\ba}{\begin{array}}
\newcommand{\ea}{\end{array}}
\newcommand{\bit}{\begin{itemize}}
\newcommand{\eit}{\end{itemize}}
\newcommand{\ben}{\begin{enumerate}}
\newcommand{\een}{\end{enumerate}}

\newcommand{\calI}{{\cal I}}

\newcommand{\calS}{{\cal S}}

\newcommand{\ie}{\emph{i.e.,}\xspace}

\newcommand{\M}[1]{{\bm #1}} %
\renewcommand{\boldsymbol}[1]{{\bm #1}}

\newcommand{\hiddenText}{{\color{gray} hidden text.}}
\newcommand{\hideWithText}[1]{\hiddenText}

\newcommand{\Real}[1]{ { {\mathbb R}^{#1} } }
\newcommand{\reals}{\Real{}}

\newcommand{\SEtwo}[1]{\ensuremath{\mathrm{SE}(2)}\xspace}
\newcommand{\SE}[1]{\ensuremath{\mathrm{SE}(#1)}\xspace}

\newcommand{\SLthree}{\ensuremath{\mathrm{SL}(3)}\xspace}
\newcommand{\SLfour}{\ensuremath{\mathrm{SL}(4)}\xspace}

\newcommand{\Simthree}{\ensuremath{\mathrm{Sim}(3)}\xspace}
\newcommand{\SEthree}[1]{\ensuremath{\mathrm{SE}(3)}\xspace}

\newcommand{\slfour}{\ensuremath{\mathfrak{sl}(4)}\xspace}
\newcommand{\intexpmap}[1]{\mathrm{Exp}\left(#1\right)}
\newcommand{\intlogmap}[1]{\mathrm{Log}\left(#1\right)}

\newcommand{\MB}{\M{B}}

\newcommand{\ME}{\M{E}}

\newcommand{\blue}[1]{{\color{blue}#1}}

\newcommand{\linkToPdf}[1]{\href{#1}{\blue{(pdf)}}}
\newcommand{\linkToPpt}[1]{\href{#1}{\blue{(ppt)}}}
\newcommand{\linkToCode}[1]{\href{#1}{\blue{(code)}}}
\newcommand{\linkToWeb}[1]{\href{#1}{\blue{(web)}}}
\newcommand{\linkToVideo}[1]{\href{#1}{\blue{(video)}}}
\newcommand{\linkToMedia}[1]{\href{#1}{\blue{(media)}}}
\newcommand{\award}[1]{\xspace} %

\newcommand{\name}{VGGT-SLAM\xspace}
\newcommand{\dr}{DUSt3R\xspace}
\newcommand{\mr}{MASt3R\xspace}

\newcommand{\wsize}{w}
\newcommand{\wfull}{\bar{w}}
\newcommand{\image}{\mathbf{M}}
\newcommand{\images}{\mathcal{I}}
\newcommand{\depth}{\mathbf{D}}
\newcommand{\conf}{\mathbf{C}}
\newcommand{\depths}{\mathcal{D}}
\newcommand{\confs}{\mathcal{C}}
\newcommand{\submap}{\mathcal{S}}
\newcommand{\sublocal}{\mathrm{latest}}
\newcommand{\subprior}{\mathrm{prior}}
\newcommand{\subloop}{\mathrm{loop}}

\newcommand{\thrdisparity}{\tau_\text{disparity}}

\newcommand{\thrnumloop}{{w}_\text{loop}}
\newcommand{\thrinterval}{\tau_{\text{interval}}}
\newcommand{\thrdesc}{\tau_{\text{desc}}}
\newcommand{\thrpoint}{\tau_{\text{conf}}}

\newcommand{\projTf}{\mathbf{H}} %
\newcommand{\vecstate}{{\boldsymbol{\xi}}}
\newcommand{\vecdelta}{{\boldsymbol{\delta}}}
\newcommand{\measfunc}[2]{h\big(#1, #2\big)}
\newcommand{\jacob}{\mathbf{J}}
\newcommand{\basisMat}{\mathbf{B}}
\newcommand{\ith}{i}
\newcommand{\jth}{j}
\newcommand{\constraintH}{\projTf^{\ith}_\jth}
\newcommand{\setHomography}{\mathcal{H}}
\newcommand{\adjointMap}{\mathrm{Ad}}
\newcommand{\genElem}{\ME}
\newcommand{\genDiag}{\MB}
\newcommand{\generator}{\mathbf{G}}
\newcommand{\suba}{i}
\newcommand{\subb}{j}

\Crefname{section}{Sec.}{Secs.} 
\Crefname{figure}{Fig.}{Figs.} 
\Crefname{table}{Table}{Tables} 
\crefname{equation}{}{}
\Crefname{equation}{}{}

\definecolor{myemerald}{rgb}{0.753, 0.898, 0.804}
\definecolor{mylightgreen}{rgb}{0.894, 0.933, 0.745}
\definecolor{myyellow}{rgb}{0.996, 0.972, 0.780}
\newcommand{\firstc}{\cellcolor{myemerald!100}}
\newcommand{\secondc}{\cellcolor{mylightgreen!100}}

\title{\name: Dense RGB SLAM \\ Optimized on the SL(4) Manifold}

\author{
  Dominic Maggio\thanks{Equal contribution.} \quad
  Hyungtae Lim\footnotemark[1] \quad
  Luca Carlone \\
  Massachusetts Institute of Technology \\
  \texttt{\{drmaggio, shapelim, lcarlone\}@mit.edu}
}

\begin{document}

\maketitle

\begin{abstract}
    We present \name, a dense RGB SLAM system constructed by incrementally and globally aligning submaps created 
    from the feed-forward scene reconstruction approach VGGT using only uncalibrated monocular cameras. 
    While related works align submaps using similarity transforms (\ie translation, rotation, and scale), 
    we show that such approaches are inadequate in the case of uncalibrated cameras. In particular, 
    we revisit the idea of 
    reconstruction ambiguity, where given a set of uncalibrated cameras with no assumption on the camera motion or scene structure, 
    the scene can only be reconstructed up to a 15-degrees-of-freedom projective transformation of the true geometry. 
    This inspires us to recover a consistent scene reconstruction 
    across submaps by optimizing over the $\SLfour$ manifold, thus estimating 15-degrees-of-freedom 
    homography transforms between sequential submaps while accounting for potential loop closure constraints. 
    As verified by extensive experiments, we demonstrate that \name achieves improved map quality using long video sequences 
    that are infeasible for VGGT due to its high GPU requirements.
\end{abstract}

\section{Introduction}

One of the most fundamental tasks in computer vision is that of simultaneous localization and mapping (SLAM) where 
given multiple monocular (or stereo) images, the task is to generate a 3D reconstruction of the scene and estimate the 6-degrees-of-freedom (DOF) pose of the cameras. 
Most approaches for this have traditionally leveraged classical multi-view geometry constraints~\cite{Nister04pami-5pt,Hartley97pami-8pt,Zheng2013ICCV-revisitPnP,Zheng14cvpr-pnpf}, data association~\cite{Lucas81,Lowe04ijcv}, and 
backend optimization such as bundle adjustment~\cite{Qin19arxiv-VINS-Fusion-poseEstimation, Qin18tro-vinsmono, Rosinol21ijrr-Kimera, Abate23iser-Kimera2, Schonberger16cvpr-SfMRevisited, Schonberger16eccv-pixelwise}.
Recently, a new paradigm of using simpler, feed-forward networks, which produce point clouds from uncalibrated input images, has 
gained increasing popularity. In this thrust, the seminal work DUSt3R~\cite{Wang24cvpr-DUST3R} takes in a pair of images and estimates dense point clouds 
of both frames in the reference frame of the first camera, thus creating a dense scene reconstruction and allowing the camera poses 
estimated easily with a 3-point RANSAC solver~\cite{Nister04cvpr,Matas05iccv}. 

To extend feed-forward reconstruction to multiple frames, VGGT~(Visual Geometry Grounded Transformer)~\cite{Wang25arxiv-VGGT} takes in an arbitrary number of frames, 
and in addition to estimating dense point clouds of each camera frame, also estimates depth maps, 
feature tracks, and camera poses and intrinsics. 
However, VGGT is limited in the number of frames that can be processed by GPU memory.
For example, in the case of an NVIDIA GeForce RTX 4090 with 24\,GB, this is limited to approximately 60 frames, 
making larger reconstructions requiring hundreds or thousands of frames infeasible. 

One may suspect that a simple, trivial solution would be to create multiple submaps with VGGT where each submap contains at least one overlapping frame, 
and solve for the scale parameter between submaps (as the reconstruction does not capture metric scale), with VGGT's estimated poses 
being used to align rotation and translation (\ie estimating a \Simthree transformation between submaps). 
While we demonstrate \Simthree optimization shows impressive reconstructions in many cases, we empirically observe that the feed-forward nature of VGGT 
with uncalibrated cameras introduces a \textit{projective ambiguity}, which in addition to the \Simthree DOF includes shear, stretch, and 
perspective DOF,
especially when the disparity between frames becomes small. This ambiguity cannot be fully resolved through a similarity transformation alone.

\begin{figure}
    \includegraphics[width=\linewidth]{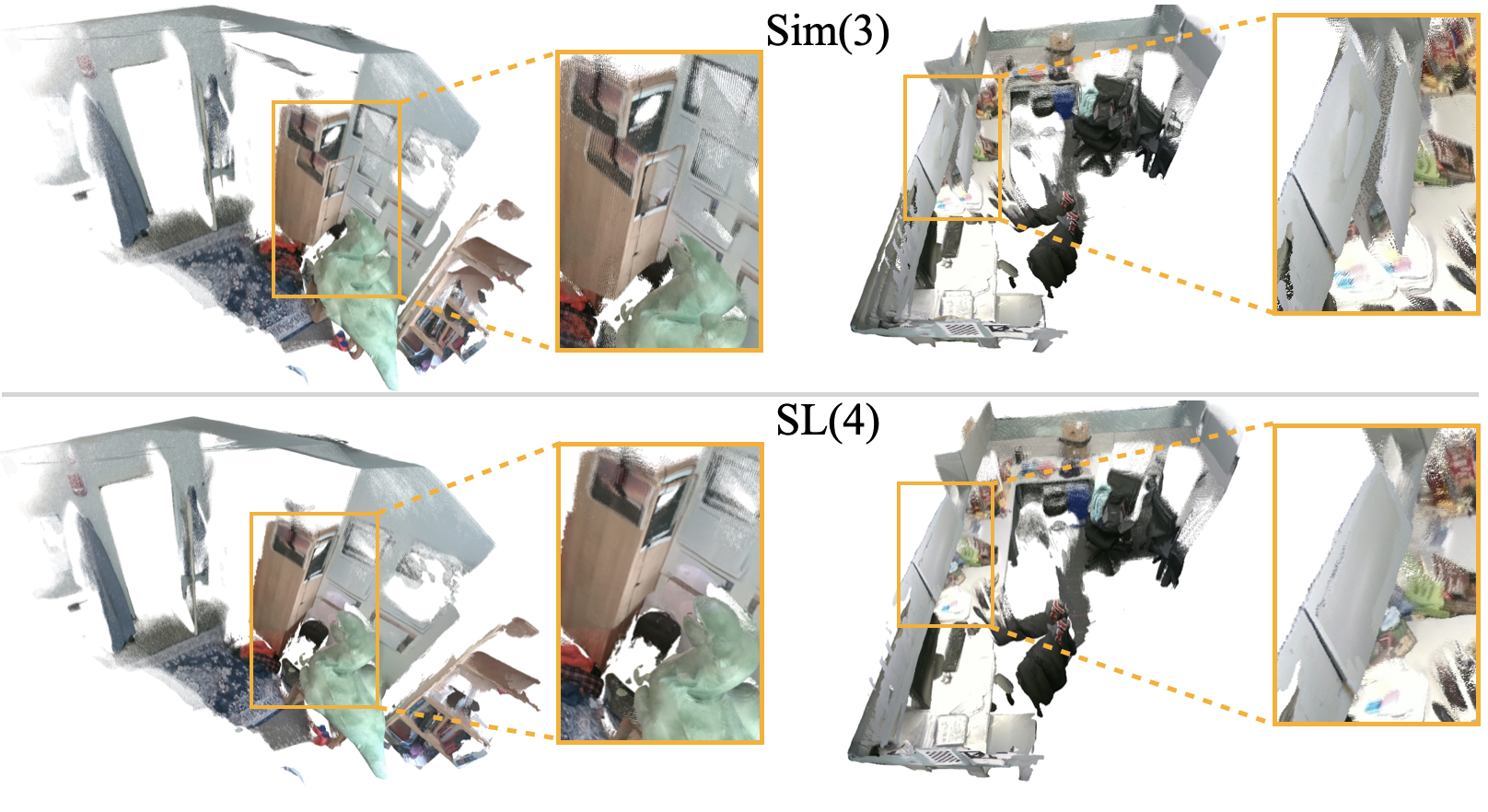}
    \vspace{-2em}
    \caption{\name alignment of 6 submaps created from VGGT using \Simthree alignment (top) and \SLfour alignment (bottom). Here, 
    \Simthree is insufficient to align submaps due to a projective ambiguity, motivating our \SLfour-based SLAM. 
    Experiments performed on a segment of the Clio~\cite{Maggio24ral-clio} apartment and cubicle scenes.}
    \label{fig:alignment}
    \vspace{-1.0em}
\end{figure}

Unveiling why a similarity transformation is sometimes insufficient for this recent transformer-based scene reconstruction method 
causes us to return to classical computer vision for answers, specifically the notion of projective ambiguity. To rectify 
a projective reconstruction to a metric reconstruction requires computing a $4\times 4$ homography matrix~\cite{Hartley04book} which belongs to the Special Linear, $\SLfour$, 
Lie group. Since this is a Lie group, we can formulate the submap alignment problem as a factor graph optimized 
on the $\SLfour$ manifold to globally align an arbitrary number of submaps given both estimates of relative homographies between sequential submaps 
and added constraints from detected loop closures. 

\paragraph{Contributions}
Firstly, we present the first SLAM system that leverages the feed-forward scene reconstruction capabilities of VGGT~\cite{Wang25arxiv-VGGT},
extending it to large-scale scenes that cannot be reconstructed from a single inference of VGGT.
Our system operates entirely with monocular RGB cameras and does not require known camera intrinsics or consistent calibration across frames.
Importantly, it achieves this without any additional training.

Secondly, while \Simthree optimization is often sufficient, we identify and analyze scenarios where projective ambiguity arises, as presented in \Cref{fig:alignment}. 
In these cases, conventional similarity transforms do not fully resolve scale and alignment issues.
We highlight this limitation and demonstrate how incorporating projective constraints addresses the problem.

Finally, we propose the first factor graph formulation that operates directly on the $\SLfour$ manifold to address projective ambiguity.
Even in practical scenarios, where projective ambiguity is less dominant, we show that $\SLfour$-based optimization achieves performance competitive with or superior to other state-of-the-art learning-based SLAM approaches,
offering a principled framework for handling cases where similarity transformations are insufficient.

\section{Related Work}

\paragraph{Classical Scene Reconstruction}

Classical scene reconstruction methods typically rely on geometric features to 
estimate camera poses and reconstruct 3D scenes from multi-view images~\cite{Mouragnon06icra,Bradley08cvpr,Furukawa10cvpr,Cramariuc22ral-Maplab, Rosinol20icra-Kimera} using~\cite{Koenderink91,Weng92pami-trifocalTensor} 
bundle adjustment~\cite{Schonberger16cvpr-SfMRevisited, Pan24eccv-GLOMAP}, by performing 
sparse feature extraction, matching, and robust pose estimation by optimizing for \SE{3} transformations. 
Several works have also performed dense SLAM~\cite{Jatavallabhula20icra-gradSlam, Whelan16ijrr-elasticFusion, Matsuki24cvpr-GSSLAM,Yan24cvpr-GSSLAM}, 
and~\cite{Cadena16tro-SLAMsurvey} provides a survey on SLAM.  
Recently, Sim-Sync~\cite{Yu24ral-SimSync} introduced a certifiably optimal algorithm
that jointly estimates camera poses and per-image scaling factors that leverage pretrained monocular depth predictions.

\paragraph{Feed-forward Scene Reconstruction}
The seminal work \dr~\cite{Wang24cvpr-DUST3R} has spawned multiple followup works in feed-forward scene reconstruction.
\dr takes in a pair of images and for each image, outputs a dense point map in the reference frame of the first camera.
From the point maps, the camera focal lengths can be estimated using the Weiszfeld algorithm~\cite{Plastria11springer-weiszfeld} 
and poses can be recovered using multiple methods such as a 3-point RANSAC~\cite{Nister04cvpr,Matas05iccv}. 
\mr follows a similar design but also outputs descriptors that can be used to generate pairwise correspondences 
between the two frames. \mr-SFM~\cite{Duisterhof24arxiv-MAST3RSfM} demonstrates global optimization of multiple images using \mr but computation scales 
quickly with the number of frames.

To extend the idea of \dr to multiple frames, Spann3R~\cite{Wang24arxiv-Spann3R} leverages a learned memory module and Cut3R~\cite{Wang25arxiv-Cut3R} uses a recurrent state 
model. Both can incrementally reconstruct a scene using multiple images, but are each limited to short sequences. 
Recently, Pow3R~\cite{Jang25arxiv-Pow3R} extends the \dr framework to optionally take in any estimates of any combination of 
camera intrinsics, poses, and depth~(which may be sparse or dense) and demonstrates substantial improvement in scene reconstruction 
and pose estimation given the added inputs.
Splatt3R~\cite{Smart24arxiv-Splatt3R} extends the \dr idea to Gaussian Splatting~\cite{Kerbl23sigraph-GS} by directly outputting the Gaussian Splatting 
parameters given two views, and PreF3R~\cite{Chen24arxiv-Pref3R} extends this to multiple views using a similar memory framework as Spann3R. 
Reloc3r~\cite{Dong24arxiv-Reloc3R} modifies the \dr framework for directly outputs relative camera poses and uses motion averaging to recover absolute 
poses with respect to a map database.

Most similar to ours is \mr-SLAM~\cite{Murai24arxiv-Mast3RSLAM}. 
\mr-SLAM leverages \mr to construct an impressive real time dense monocular SLAM system that does 
not require known calibration. Their pipeline also includes efficient optimization over $\Simthree$ poses and loop closures. 
Since \mr is limited to two input frames at a time, here, we desire to build on top of the more powerful VGGT architecture for a SLAM 
system which can leverage broader information of the scene by taking in an arbitrary number of frames for feed-forward reconstruction (bounded by computational limits) 
and provides direct estimates of camera poses. 
However, as mentioned, fusing submaps from VGGT goes beyond a traditional 
point cloud registration problem as alignment cannot be effectively performed with only a similarity transformation.
Unlike \mr-SLAM, as will be discussed in \Cref{sec:local_align}, we do not need to estimate correspondences between frames. 

An alternative paradigm, scene coordinate regression, with works such as ACE~\cite{Brachmann23cvpr-ace} and DSAC\text{*}~\cite{Brachmann21cvpr-dsacstar}, estimates world points from images 
with respect to a global scene frame by using a scene specific trained network.

\paragraph{Optimization over the Special Linear group}
To the best of our knowledge, we are the first work to create a factor graph optimization for point cloud alignment on the \SLfour 
manifold. Prior works use optimization on the \SLthree manifold (corresponding to the 
8-DOF homography matrix commonly used in image 
alignment) for aligning multiple images for panoramic 
stitching~\cite{Hamel11cdc-sl3Stiching, Shum00ijcv-panoramicAlignment, Mei06iros, Mei06bmvc, Mei08tro, Lovegrove10eccv-panoramicSLAM} 
and dense SLAM~\cite{Lovegrove12icl-parametric}. 
The 15-DOF homography matrix is used for 
classical tasks such as auto-calibration~\cite{Hartley94cvpr-autocalibration}, and good practices for estimating  
homography are extensively studied in~\cite{Hartley00}.

\section{Review: VGGT}
Here, we provide the relevant preliminaries of VGGT~\cite{Wang25arxiv-VGGT}. VGGT takes as input an 
image set $\images = \{\image_1, \cdots , \image_{\wfull}\}$, which consists of $\wfull$ images, 
tokenizes them with a fine-tuned DINO~\cite{Oquab23arxiv-dinov2} backbone, and then 
applies Alternating-Attention (alternating between applying global and frame-wise attention). The output tokens can 
then be passed to a camera head to estimate intrinsics and camera poses (defined with respect to the first frame), 
or to Dense Prediction Transformer~(DPT) heads~\cite{Ranftl21iccv-DPT},
which outputs dense depth maps for each image, 
a dense point map (where the points of each camera are defined with respect to the first camera), 
and dense features for point tracking, with confidence estimates provided for each. 

In this paper, we use the dense depth maps $\depths = \{\depth_1, \cdots, \depth_{\wfull}\}$ and confidence score 
maps $\confs = \{\conf_1, \cdots, \conf_{\wfull}\}$ (as they are fully dense, the width and height of corresponding components in $\image$, $\depth$, and $\conf$ are the same). 
We refer to the outputs from each $\images$ as submap $\submap$, which will correspond to a node in pose graph optimization for \name.
We do not 
run the 3D point DPT head as it was observed in~Wang \textit{et al.}~\cite{Wang25arxiv-VGGT} 
that more accurate point clouds can be achieved by inverse projecting $\depth$ using projection matrices from the camera head, giving 
us a dense point cloud which is defined with respect to the coordinate frame 
of the first camera in $\images$. We denote this point cloud as $\textbf{X}^\submap$.
To filter unreliable points, 
we prune points whose associated confidence values in the confidence maps are less than $\thrpoint$ of the average confidence across $\confs$.

\section{\name}
\label{sec:method}

Here we describe the design of our \name system. In \Cref{sec:submap} we determine how to generate a list of frames that will be 
passed to VGGT to produce a local submap, $\calS$. In \Cref{sec:local_align}, we provide a discussion of projective 
ambiguity and show how we can align two overlapping submaps by 
estimating a relative 15-DOF homography matrix between sequential submaps, 
and in \Cref{sec:loop_closure}, we describe the process of adding loop closure constraints between non-sequential submaps.
Finally, in \Cref{sec:backend} we show how we can globally optimize all submap alignments into a consistent map 
by optimizing on the $\SLfour$ manifold.

\subsection{Incremental submap-based keyframe selection and generation}
\label{sec:submap}

First, we begin by describing how to incrementally construct submaps and organize keyframes within each submap from sequentially incoming images.
For this, we construct an image set $\images_\sublocal$. 
As is typical in visual SLAM~\cite{Rosinol21ijrr-Kimera, Murai24arxiv-Mast3RSLAM,Song22ral-DynaVINS},
we select an image as a keyframe 
if disparity (which we estimate using Lucas-Kanade~\cite{Lucas81}) with respect to the previous keyframe is larger 
than a user-defined threshold $\thrdisparity$. 
Even though VGGT demonstrates monocular depth capabilities~\cite{Wang25arxiv-VGGT} from learned priors,
utilizing frames with sufficient disparity improves relative depth estimation performance 
as it adds multi-view information and additionally reduces the number of frames to process.

If sufficiently high disparity is estimated, the current frame is designated a keyframe and added 
to a list of frames, $\images_\sublocal$, until the size of the list reaches a set limit $\wsize$. 
In addition to $\images_\sublocal$, each submap's associated image set is constructed by concatenating two additional sets of frames. 
The first set includes a single frames chosen as the last non-loop-closure frame from the previous submap, denoted as $\image_\subprior$.
Up to $\thrnumloop$ frames to be used for loop closures (discussed in \Cref{sec:loop_closure}) may also be appended at the end of the collection,
forming the final image set for the submap as $\images_\sublocal \leftarrow \{\image_\subprior\} \cup \mathcal{I}_\sublocal \cup \images_\subloop$. 
This image set is then passed to VGGT to generate the submap, $\calS_\sublocal$.

\subsection{Local submap alignment addressing projective ambiguity}
\label{sec:local_align}

Given two overlapping submaps $\calS_i$ and $\calS_j$ generated as described in \Cref{sec:submap}, 
which have point clouds $\textbf{X}^{\calS_i}$ and $\textbf{X}^{\calS_j}$ in their respective local submap frames,
our objective is to solve for a transformation, $\textbf{H}^i_j\in \mathbb{R}^{4\times4}$ that aligns the two 
submaps such that for any noise-free corresponding points $\textbf{X}^{\calS_i}_a,~\textbf{X}^{\calS_j}_b~\in~\reals^{3}$, the following relation holds:

\begin{equation}
    \label{eq:homography}
    \textbf{X}^{\calS_i}_a = \textbf{H}^i_j \textbf{X}^{\calS_j}_b, 
\end{equation}

where for simplicity, we overload notation such that $\textbf{X}^{\calS}$ is in homogeneous coordinates when multiplied by the $4\times4$ homography matrix. 
Under a typical 3D point cloud alignment problem, for example from LIDAR SLAM~\cite{Lim24ijrr-Quatropp}, $H$ would represent a translation and rotation 
in \SE{3}. If the point clouds additionally differ in scale, then $\textbf{H}$ would be on \Simthree, the group of similarity transformations. 
However, here we do not have typical 
point clouds as $\textbf{X}^{\calS}$ is constructed by uncalibrated cameras. Thus, we 
recall \emph{the Projective Reconstruction Theorem~\cite[Chapter 10.3]{Hartley04book}, which in summary states that 
if correspondences between two images from uncalibrated cameras uniquely determine the fundamental matrix, then 
the correspondences may be used to reconstruct the corresponding 3D points 
up to a 15-DOF homography transformation. 
This transform is the same for any such corresponding points, except those on the line connecting the camera 
centers as these points cannot be reconstructed uniquely.} 
Relevant to our setup, the Projective Reconstruction 
Theorem also applies to a reconstruction with more than two cameras~\cite{Hartley04book}. 
Thus, in the most general case, 
the reconstruction computed using a set of uncalibrated cameras differs from a metrically correct reconstruction by a 
projective transformation (\ie homography) \textbf{H}.
The matrix \textbf{H} has 
15 DOF and belongs to the \textit{special linear group}, \SLfour. 
The \SLfour~group consists of all real-valued $4 \times 4$ matrices with unit determinant.
Note this is not the same as the more common 8 DOF 
homography matrix commonly used in planar computer vision tasks such as image warping, which belongs to \SLthree. 
The reconstruction can be transformed to an affine reconstruction (\ie parallel lines are preserved) when scene priors 
are available, for example if points are known to lie on parallel 
lines. If further priors are known, such as lines in the scene are orthogonal, then the reconstruction can be converted to 
a metric reconstruction (differing 
by only a similarity transform to the true Euclidean reconstruction). VGGT is thus able to leverage learned scene priors to potentially estimate 
metric reconstruction, but as we have shown in \Cref{fig:alignment}, in the most general case when estimates of scene priors are 
unreliable, the reconstruction differs by a projective ambiguity, 
requiring a 15-DOF homography matrix to rectify. We will now estimate such a homography. 

By our construction of the submaps that they share a same frame, 
we have an atypical advantage in solving for $\textbf{H}$ as we have a dense set of correspondences 
without needing to estimate associations.

As is well known by the direct reconstruction method~\cite{Hartley04book}, the optimal homography in \eqref{eq:homography} can be solved in 
closed form as a solution to the following homogeneous linear system:

\begin{equation}
    \label{eq:homography_solve}
    \textbf{A}_k \textbf{h} = 0
\end{equation}

with $\textbf{h} \in \reals^{16}$ containing the flattened parameters of the homography and $\textbf{A}_k$ contains constraints for a particular 
pair of 3D points. A minimum solution requires 5 points (\ie $k\in\{1:5\}$), and to build in robustness 
to incorrect depth measurements from VGGT, we solve \Cref{eq:homography_solve} using RANSAC~\cite{Fischler81} with a 5-point solver. 
As the homography matrix is estimated up to scale, we scale by the fourth root of the determinant such that the determinant is unity 
and the resulting matrix belongs to \SLfour.

\paragraph{Transformation of camera poses via homography}
Using $\textbf{H}^i_j$, the camera poses can be corrected using the following~\cite{Hartley04book}: $\textbf{P}_i = {\textbf{H}^i_j}^{-1} \textbf{P}_j$, 
where $\textbf{P}\in \mathbb{R}^{4\times 4}$ is the camera matrix created from the poses and intrinsic estimates from VGGT. We can then decompose $\textbf{P}$ to 
recover the camera pose.

\subsection{Loop closures}
\label{sec:loop_closure}

Our procedure for creating loop closures for \name consists of two steps: (i)~performing image retrieval~(\ie~setting $\calI_\subloop$ in \Cref{sec:submap}),
and (ii)~estimating relative homographies, which are then added to the factor graph as loop closure constraints (\Cref{sec:backend}).
First, for image retrieval, when constructing a submap, we compute and store an image descriptor for each keyframe using SALAD~\cite{Izquierdo24cvpr-SALAD}.
Then, once $\images_\sublocal$ reaches its size threshold $\wsize$, 
we search over the image descriptors in the previous submaps~$\calS_i \; \forall i \in \{1:\text{lastest}-\thrinterval\}$ 
to fetch a set of frames of size $\thrnumloop$ that have the highest similarity~(using the L2 norm) 
to any of the keyframes in $\images_\sublocal$, and also exceed a user-defined similarity threshold $\thrdesc$ to reduce false positive matches. 
These frames make up $\calI_\subloop$, which is added to the list of keyframes for the current submap, 
and then all frames are sent to VGGT as described in \Cref{sec:submap}.

Next, given the estimated submap, $\calS_\sublocal$, from VGGT, we estimate the relative homographies between 
the loop closure frames in $\calS_\sublocal$ and the submaps, $\calS_i$, retrieved during the image retrieval process described above. 
As in \Cref{sec:local_align}, we again have the benefit of not requiring an estimate of correspondences to compute 
homographies for loop closures; thus, we can directly use \eqref{eq:homography_solve} between the frames in $\calI_\subloop$ and 
their respective identical frames in the submap where they originated. This then provides $\thrnumloop$ loop closure 
constraints between $\calS_\sublocal$ and the corresponding submaps.

Note that a potential alternative is to get the descriptor using the output tokens from VGGT's fine-tuned DINO backbone. 
This alleviates 
using a separate descriptor module and storing the physical images. However, this requires storing larger features in memory compared to 
the relatively small SALAD features, and the system memory needed to store the images in our base approach is relatively low.

\subsection{Backend: Nonlinear factor graph optimization on the \SLfour manifold}
\label{sec:backend}

Given all relative homographies $\constraintH$ between submaps $\mathcal{S}_i$ and $\mathcal{S}_j$,
our goal is to compute the absolute homographies $\textbf{H}_i$ that transform all submaps into a common global reconstruction.
To achieve this, we formulate a nonlinear factor graph optimization problem\footnote{In our case, this is an extension of pose graph optimization, where we estimate absolute poses from pairwise pose measurements.} based on Maximum A Posteriori~(MAP) estimation~\cite{blanco10personal-se3,Ebadi24tro-surveySLAMSubt,Grisetti10tutorial}.
Specifically, we estimate the absolute homographies by minimizing the following cost function under Gaussian noise assumptions on the relative homographies:

\begin{equation}
    \label{eq:sl4_optim}
    \hat{\setHomography}=\underset{\projTf \in \SLfour}{\operatorname{argmin}} \sum_{(\ith, \jth) \in \mathcal{L}} \left\| \intlogmap{\projTf^{-1}_\ith \projTf_\jth \left(\constraintH\right)^{-1}} \right\|_{\Omega^\projTf_{\ith \jth} }^{2},
\end{equation}
where $\mathrm{Log}(\cdot)$ is the mapping function that transforms a group element to a (vectorized) element of the corresponding Lie algebra,
$\mathcal{L}$ denotes an index set of constraints that includes odometry and loop closures,
and we set $\Omega^\projTf_{\ith \jth} \in \mathbb{R}^{15 \times 15}$ to the identity matrix.

To solve \eqref{eq:sl4_optim}, we iteratively compute state increments by solving a linearized least squares problem.
To this end, we define $\vecstate \in \mathbb{R}^{15}$ as the tangent-space parameterization of \SLfour,
the mapping function $\mathrm{Exp} : \mathbb{R}^{15} \rightarrow \SLfour$, which satisfies $\mathrm{Log}\big(\mathrm{Exp}\left(\vecstate \right)\big) = \vecstate$ and $\intexpmap{\vecstate} = \mathrm{exp}(\vecstate^\wedge) = \projTf$.
In particular, $\vecstate^\wedge$ is a Lie algebra element of \slfour, computed by summing the $k$-th component of $\vecstate$ with its $k$-th corresponding generator $\generator_k \; \forall k:\{1:15\}$~(\ie~$\vecstate^\wedge = \sum_{k=1}^{15} \vecstate_k \, \generator_k$)~\cite{Eade13liegroups}. 
More details can be found in \Cref{appendix:tangent_space}. 

Next, defining the measurement function as $\measfunc{\vecstate_\suba}{\vecstate_\subb} = \intlogmap{\projTf_\suba^{-1}\projTf_\subb}$, 
the incremental update of each pose can be approximated using Taylor's expansion~\cite{Dellaert12tr} as follows:

\begin{equation}
  \measfunc{\vecstate_{\suba} \oplus \vecdelta_\suba}{\vecstate_{\subb}\oplus \vecdelta_\subb} \simeq \measfunc{\vecstate_{\suba}}{\vecstate_{\subb}} \oplus\left\{\jacob_\suba \vecdelta_\suba + \jacob_\subb \vecdelta_\subb \right\}, \; \vecstate \oplus \vecdelta = \projTf \, \intexpmap{\vecdelta}
\end{equation}

where $\jacob_\suba = -\adjointMap_{\projTf_\suba^{-1}\projTf_\subb}$ and $\jacob_\subb = I_{15\times15}$.
Here, $\adjointMap_{\projTf}$ is the adjoint map, defined as $\adjointMap_{\projTf} = \basisMat^{-1}{\projTf} \otimes {\projTf}^{-\intercal}\basisMat$~\cite{Eade13liegroups},
where $\basisMat=\left[\operatorname{vec}\left(\generator_1\right) \;  \operatorname{vec}\left(\generator_2\right) \; \cdots \; \operatorname{vec}\left(\generator_{15}\right) \right] \in \mathbb{R}^{16 \times 15}$ in the case of the \SLfour manifold~\cite{Conder92presentations},
and $\otimes$ denotes the Kronecker product, which forms a block matrix by multiplying each element of the first matrix with the entire second matrix.

Finally, we can formulate the linearized residuals and the resulting local problem at the linearization point $\projTf^\ith_\jth$ as follows:

\begin{equation}
    \label{eq:linearized}
    \hat{\mathcal{D}}=\underset{\vecdelta \in \mathcal{D}}{\operatorname{argmin}} \sum_{(\ith, \jth) \in \mathcal{L}} \left\| \boldsymbol{e}_{\ith \jth} + \jacob_\suba \vecdelta_\suba + \jacob_\subb \vecdelta_\subb \right\|_{\Omega^\projTf_{\ith \jth} }^{2},
    \; \boldsymbol{e}_{\ith \jth} = \intlogmap{\projTf^{-1}_\ith \projTf_\jth \left(\constraintH\right)^{-1}}. 
\end{equation}
To solve \eqref{eq:linearized}, we use the Levenberg-Marquardt optimizer~\cite{Rosen12icra}, and at each iteration, the poses are updated on the Lie group as $\projTf \leftarrow \projTf \, \mathrm{Exp}(\hat{\vecdelta})$~\cite{Sola17Quaternion}.

\section{Experiments}
\label{sec:experiments}

We follow similar experiments as \mr-SLAM to evaluate camera pose estimation and dense reconstruction in \Cref{sec:pose} and 
\Cref{sec:reconstruction} respectively, demonstrate qualitative results in \Cref{sec:qualitative}, and finally perform 
ablations in \Cref{sec:ablations}.

\subsection{Experimental setup}

We evaluate \name on standard RGB SLAM benchmarks to assess both camera pose estimation accuracy and dense mapping quality.
For evaluation of pose estimation,
we employ the 7-Scenes~\cite{Shotton13cvpr} and TUM RGB-D~\cite{Sturm12iros-TUM-RGB-D} datasets,
and report root mean square error~(RMSE) of the absolute trajectory error~(ATE) using evo~\cite{Grupp17evo}.
Since 7-Scenes~\cite{Shotton13cvpr} provides scene ground truth, 
this dataset is also used to evaluate dense mapping quality in terms of \textit{accuracy}, \textit{completion}, and \textit{Chamfer distance}~\cite{Murai24arxiv-Mast3RSLAM}. 

As baseline approaches, we primarily compare \name with DROID-SLAM~\cite{Teed21nips-DROID-SLAM} and \mr-SLAM~\cite{Murai24arxiv-Mast3RSLAM} as the state-of-the-art 
learning-based SLAM approaches in uncalibrated scenarios (and Spann3R~\cite{Wang24arxiv-Spann3R} for dense evaluation). 
We use reported numbers from \mr-SLAM~\cite{Murai24arxiv-Mast3RSLAM} for baselines, except for the uncalibrated version of DROID-SLAM.
Although DROID-SLAM requires camera intrinsics, we also evaluate it in an uncalibrated setting by estimating intrinsics with an automatic calibration pipeline~\cite{Veicht24eccv-Geocalib}, as is suggested by Murai~\textit{et al.}~\cite{Murai24arxiv-Mast3RSLAM}.
While our approach operates without camera calibration, we also include comparison with state-of-the-art methods~\cite{Campos21-TRO,Lipson24eccv-DeepPatch,Zhang23iccv-GOSLAM,Teed20iclr-DEEPV2D,Czarnowski20ral-Deepfactors,Zhu243dv-NICERSLAM} provided with camera intrinsics. 
Due to potential randomness in our approach caused by RANSAC, we report the average performance over five runs, which have a low spread (small standard deviation) as 
shown in \Cref{sec:ablations}. 

We refer to a simpler \Simthree version of \name as Ours (\Simthree), for which we follow similar structure as our 
\SLfour pipeline except we align relative rotation and translation between submaps using pose estimates from VGGT and estimate a 
scale correction by comparing the estimated point clouds of the overlapping frames. 
Loop closures and relative factors are added to the factor graph as \SE{3} factors. 

We use an NVIDIA GeForce RTX 4090 GPU with AMD Ryzen Threadripper 7960X CPU.
For parameters, we set
$\thrnumloop = 1$, $\thrdisparity = 25$ pixels, 
$\thrinterval = 2$,
$\tau_\text{desc} = 0.8$, 
and 
$\thrpoint$ = 25\%. We also use 300 RANSAC iterations with a threshold of 0.01. We show evaluations of 
both the \SLfour and \Simthree version 
of \name with different submap sizes (\ie different values for $w$).

\subsection{Pose estimation evaluation}
\label{sec:pose}

As shown in Tables~\ref{tab:7scenes_ate} and \ref{tab:tum_ate}, \name performs comparable to the top performing 
uncalibrated baselines on 7-Scenes and TUM RGB-D. On 7-Scenes for instance, \name has approximately the same 
average APE as the top performing baseline \mr-SLAM. 
On the TUM dataset, the \SLfour version of \name performs the best overall with an average error of 0.053 m. 
This demonstrates that we are able to extend VGGT to multiple sequences while 
introducing a new category of SLAM system by optimizing submap alignment as \SLfour factors. Here, we observe that our \Simthree 
version also performs well, as these scenes are generally cases where VGGT is able to leverage strong priors for metric reconstruction. 
Thus, while we have shown cases where \SLfour is needed (\Cref{fig:alignment}), the addition of higher degrees of freedom with our novel 
SLAM formulation maintains competitive performance, while improving some more challenging cases.

\newcommand{\floor}{\texttt{floor}\xspace}
One particular scene where our method underperforms in on the TUM \floor scene. This highlights a challenge of estimating homography, which is 
the presence of degeneracy in 
the case of a planar scene. The floor scene contains several images that only view the flat floor leading to non-unique solutions for the homography matrix,
which causes the overall reconstruction to diverge. Building robustness for the planar case is an important 
component for \SLfour SLAM, which we leave as an exciting direction for future work. 
The TUM \texttt{360} scene is particularly challenging for smaller submap sizes (although handled well 
with $w=32$) because smaller submap are more likely to encounter approximately pure rotation in this scene, which can have reduced depth accuracy and hence a higher outlier 
ratio when running 5-point RANSAC to estimate homography.
\newcommand{\cg}{\color{gray!70}}

\begin{table}[h]
    \vspace{-0.5em}
    \centering
    \caption{Root mean square error~(RMSE) of absolute trajectory error (ATE) on 7-Scenes~\cite{Shotton13cvpr} (unit: m).
    The gray rows indicate the results using the calibrated camera intrinsics and the~*~symbol indicates that the baseline is evaluated in the uncalibrated mode.
    Green is best and light green is second best.}\label{tab:7scenes_ate}
    \scriptsize
    \begin{tabular}{l|lcccccccc} %
    \toprule
    \multirow{2}{*}{} & \multirow{2}{*}{Method} & \multicolumn{7}{c}{Sequence} &  \multirow{2}{*}{Avg}  \\ \cmidrule(lr){3-9}
                      & &\texttt{chess} &\texttt{fire} &\texttt{heads} &\texttt{office} &\texttt{pumpkin} &\texttt{kitchen} &\texttt{stairs} & \\
    \midrule
    \cg
    \parbox[t]{1mm}{\multirow{3}{*}{\rotatebox[origin=c]{90}{Calib.}}} 
                      & \cg NICER-SLAM~\cite{Zhu243dv-NICERSLAM} & \cg \textbf{0.033} & \cg 0.069 & \cg 0.042 & \cg 0.108 & \cg 0.200 &  \cg \textbf{0.039} & \cg 0.108 & \cg 0.086 \\
                      & \cg DROID-SLAM~\cite{Teed21nips-DROID-SLAM} & \cg {0.036} & \cg {0.027} & \cg {0.025} & \cg \textbf{0.066} & \cg 0.127 & \cg  {0.040} & \cg  {0.026} & \cg  {0.049} \\
                      & \cg \mr-SLAM~\cite{Murai24arxiv-Mast3RSLAM} & \cg 0.053 & \cg \textbf{0.025} &  \cg \textbf{0.015} & \cg {0.097} & \cg  \textbf{0.088} & \cg 0.041 & \cg  \textbf{0.011} & \cg  \textbf{0.047} \\ \midrule
    \parbox[t]{1mm}{\multirow{4}{*}{\rotatebox[origin=c]{90}{Uncalib.}}} 
                      &DROID-SLAM*~\cite{Teed21nips-DROID-SLAM} & 0.047 & 0.038 & 0.034 & 0.136 & 0.166 & 0.080 &  \secondc 0.044 & 0.078  \\
                      &\mr-SLAM*~\cite{Murai24arxiv-Mast3RSLAM} & 0.063 & 0.046 & 0.029 & \firstc 0.103 & \firstc {0.114} &0.074 & \firstc 0.032 & \firstc 0.066 \\
                      &Ours ($\Simthree, w=32$) & \secondc 0.037 & \firstc 0.026 & \firstc 0.018 &  0.104 & \secondc 0.133 & \secondc0.061 & 0.093 & \secondc 0.067 \\
                      &Ours ($\SLfour, w=32$) & \firstc 0.036 & \secondc 0.028 & \firstc 0.018 & \firstc 0.103 & \secondc 0.133 & \firstc0.058 & 0.093 & \secondc 0.067  \\
    \bottomrule
    \end{tabular}
\end{table}

\begin{table}[h]
    \vspace{-0.5em}
    \centering
    \caption{Root mean square error~(RMSE) of absolute trajectory error~(ATE) on TUM RGB-D~\cite{Sturm12iros-TUM-RGB-D} (unit: m).
    The gray rows indicate the results using the calibrated camera intrinsics and the * symbol indicates that the baseline is evaluated in the uncalibrated mode.
    Green is best and light green is second best.}\label{tab:tum_ate}
    \scriptsize
    \begin{tabular}{l|lcccccccccc} 
    \toprule
    \multirow{2}{*}{} & \multirow{2}{*}{Method} & \multicolumn{9}{c}{Sequence} &  \multirow{2}{*}{Avg}  \\ \cmidrule(lr){3-11}
    & &\texttt{360} &\texttt{desk} &\texttt{desk2} &\texttt{floor} &\texttt{plant} &\texttt{room } &\texttt{rpy} &\texttt{teddy} &\texttt{xyz} & \\
    \midrule
    \cg
    \parbox[t]{1mm}{\multirow{8}{*}{\rotatebox[origin=c]{90}{Calib.}}} 
    & \cg ORB-SLAM3~\cite{Campos21-TRO} & $ \cg \times$ & \cg {0.017} & \cg 0.210 &$ \cg \times$ & \cg 0.034 &$ \cg  \times$ & \cg $\times$ &  \cg $\times$ & \cg \textbf{0.009} & \cg  N/A \\
    & \cg DeepV2D~\cite{Teed20iclr-DEEPV2D} & \cg 0.243 & \cg 0.166 & \cg 0.379 & \cg 1.653 & \cg 0.203 & \cg 0.246 & \cg 0.105 & \cg 0.316 & \cg 0.064 & \cg 0.375 \\
    & \cg DeepFactors~\cite{Czarnowski20ral-Deepfactors} & \cg 0.159 & \cg 0.170 & \cg 0.253 & \cg 0.169 & \cg 0.305 & \cg 0.364 & \cg 0.043 & \cg 0.601 & \cg 0.035 & \cg 0.233 \\
    & \cg DPV-SLAM~\cite{Lipson24eccv-DeepPatch} & \cg 0.112 & \cg 0.018 & \cg 0.029 & \cg 0.057 & \cg 0.021 & \cg 0.330 & \cg 0.030 & \cg 0.084 & \cg {0.010} & \cg 0.076 \\
    & \cg DPV-SLAM++~\cite{Lipson24eccv-DeepPatch} & \cg 0.132 & \cg 0.018 & \cg 0.029 & \cg 0.050 & \cg 0.022 & \cg 0.096 & \cg 0.032 & \cg 0.098 & \cg {0.010} & \cg 0.054 \\
    & \cg GO-SLAM~\cite{Zhang23iccv-GOSLAM} & \cg 0.089 & \cg \textbf{0.016} & \cg {0.028} & \cg {0.025} & \cg 0.026 & \cg {0.052} & \cg \textbf{0.019} & \cg 0.048 & \cg {0.010} & \cg {0.035} \\
    & \cg DROID-SLAM~\cite{Teed21nips-DROID-SLAM} & \cg 0.111 & \cg 0.018 & \cg 0.042 & \cg \textbf{0.021} & \cg \textbf{0.016} & \cg \textbf{0.049} & \cg {0.026} & \cg 0.048 & \cg 0.012 & \cg 0.038 \\
    & \cg \mr-SLAM~\cite{Murai24arxiv-Mast3RSLAM} & \cg \textbf{0.049} & \cg \textbf{0.016} & \cg \textbf{0.024} & \cg {0.025} & \cg {0.020} & \cg 0.061 & \cg 0.027 & \cg \textbf{0.041} & \cg \textbf{0.009} & \cg \textbf{0.030} \\
    \midrule
    \parbox[t]{1mm}{\multirow{4}{*}{\rotatebox[origin=c]{90}{Uncalib.}}} &DROID-SLAM*~\cite{Teed21nips-DROID-SLAM} &0.202 & \secondc 0.032 &0.091 & \secondc 0.064 &0.045 &0.918 &0.056 &0.045 & \firstc 0.012 &0.158 \\
    &\mr-SLAM*~\cite{Murai24arxiv-Mast3RSLAM} &\firstc {0.070} & 0.035 & \secondc 0.055 & \firstc 0.056 &0.035 & 0.118 &\secondc0.041 &0.114 & 0.020 & \secondc 0.060 \\
    &Ours~(\Simthree, $w=32$) & 0.123 & 0.040 & \secondc 0.055 & 0.254 & \firstc 0.022 & \firstc 0.088 & \secondc0.041 & \firstc 0.032 & 0.016 & 0.074 \\
    &Ours~(\SLfour, $w=32$) & \secondc 0.071 & \firstc 0.025 & \firstc 0.040 & 0.141 & \secondc 0.023 & \secondc 0.102 & \firstc 0.030 & \secondc 0.034 & \secondc 0.014 & \firstc 0.053 \\
    \bottomrule
    \end{tabular}
    \vspace{-1.0em}
\end{table}

\subsection{Dense reconstruction evaluation}
\label{sec:reconstruction}

Following the protocol of \mr-SLAM, we provide dense reconstruction performance on 7-Scenes; see \Cref{tab:7scenes_recon}. 
Here, we observe that while performance is comparable across methods,
\name achieves the best performing accuracy and Chamfer distance, demonstrating the high 
accuracy of our dense point cloud reconstruction. 

\begin{wraptable}{r}{0.5\textwidth}
    \vspace{-1.2em}
    \centering
    \caption{Root mean square error~(RMSE) reconstruction evaluation on 7-Scenes~\cite{Shotton13cvpr} (unit: m). $@n$ indicates a keyframe every $n$ images.}
    \setlength{\tabcolsep}{2.0pt}
    \scriptsize
    \begin{tabular}{l|lcccc}
    \toprule
    \multirow{2}{*}{} & \multirow{2}{*}{Method} & \multicolumn{4}{c}{7-Scenes} \\ \cmidrule(lr){3-6}
     & & ATE\,$\downarrow$ & Acc.\,$\downarrow$ & Complet.\,$\downarrow$ & Chamfer\,$\downarrow$ \\
    \midrule
    \cg \parbox[t]{2mm}{\multirow{4}{*}{\rotatebox[origin=c]{90}{Calib.}}}
    & \cg DROID-SLAM~\cite{Teed21nips-DROID-SLAM} & \cg 0.049 & \cg 0.141 & \cg 0.048 & \cg 0.094  \\
    & \cg \mr-SLAM~\cite{Murai24arxiv-Mast3RSLAM} & \cg \textbf{0.047} & \cg 0.089 & \cg 0.085 & \cg 0.087  \\
    & \cg Spann3R @20~\cite{Wang24arxiv-Spann3R}  & \cg N/A & \cg 0.069 & \cg 0.047 & \cg 0.058 \\ 
    & \cg Spann3R @2~\cite{Wang24arxiv-Spann3R}   & \cg N/A & \cg 0.124 & \cg \textbf{0.043} & \cg 0.084 \\
    \midrule
    \parbox[t]{2mm}{\multirow{3}{*}{\rotatebox[origin=c]{90}{Uncalib.}}}
    & \mr-SLAM*~\cite{Murai24arxiv-Mast3RSLAM}  & \firstc 0.066 & \secondc 0.068 & \firstc 0.045 & \secondc0.056  \\
    & Ours (\Simthree, $w=32$) & \secondc 0.067 & \firstc \textbf{0.052} & 0.062 & 0.057 \\
    & Ours (\SLfour, $w=32$) & \secondc 0.067 & \firstc \textbf{0.052} & \secondc 0.058 & \firstc \textbf{0.055} \\
    \bottomrule
    \end{tabular}
    \vspace{-1em}
    \label{tab:7scenes_recon}
\end{wraptable}

\subsection{Qualitative results}
\label{sec:qualitative}
\begin{figure}
    \centering
    \includegraphics[height=0.3\linewidth]{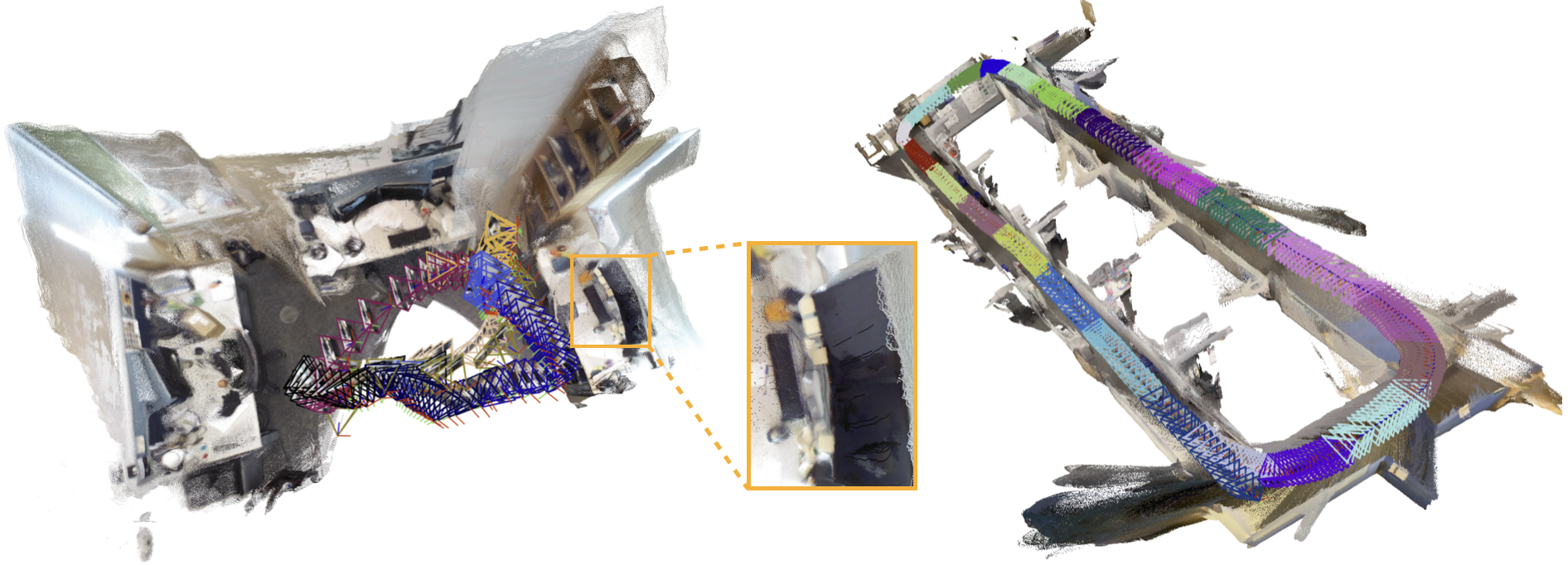}
    \caption{Reconstruction and pose estimates from \name on the office scene from 7-Scenes showing 8 submaps and from a 
    custom scene showing a 55 meter loop around an office corridor with 22 submaps. Both use $w=16$. 
    Different frame colors indicate the submap associated with each frame.}
    \label{fig:dense}
    \vspace{-2em}
\end{figure}

We present qualitative results to illustrate the mapping fidelity of \name using \SLfour optimization. 
In \Cref{fig:dense}, we show an example reconstruction from the office scene in 7-Scenes and from a longer 55 meter long 
trajectory that loops inside an office corridor. In addition to the dense reconstruction, we also 
show all mapped camera poses, where different colors indicate the submap associated with each frame. 
In particular, the office corridor loop clearly shows 22 different submaps 
which have been joined into a globally consistent map with a loop closure at the end of the trajectory. 

In \Cref{fig:alignment}, we show two select examples where using only \Simthree is unable to align overlapping submaps while 
our \SLfour alignment strategy is able to rectify the projective ambiguity between submaps. Thus, while we have shown that \Simthree 
generally achieves accurate performance across our quantitative experiments, in the general case where a feed-forward reconstruction 
method like VGGT is unable to estimate a metric reconstruction (for reasons discussed in \Cref{sec:local_align}) due to the computational limits, 
our introduction of an \SLfour-based SLAM system shows promise in leveraging the potential of a high accuracy, 
dense, learning-based SLAM system. For \Cref{fig:alignment}, $\thrdisparity$ is set to 0 to highlight the impact of projective ambiguity,
which degrades the performance of \Simthree\ alignment and affects overall map quality.

\subsection{Ablations}
\label{sec:ablations}

\begin{figure}[h]
    \centering
    \begin{subfigure}[t]{1.0\linewidth}
        \includegraphics[width=\linewidth]{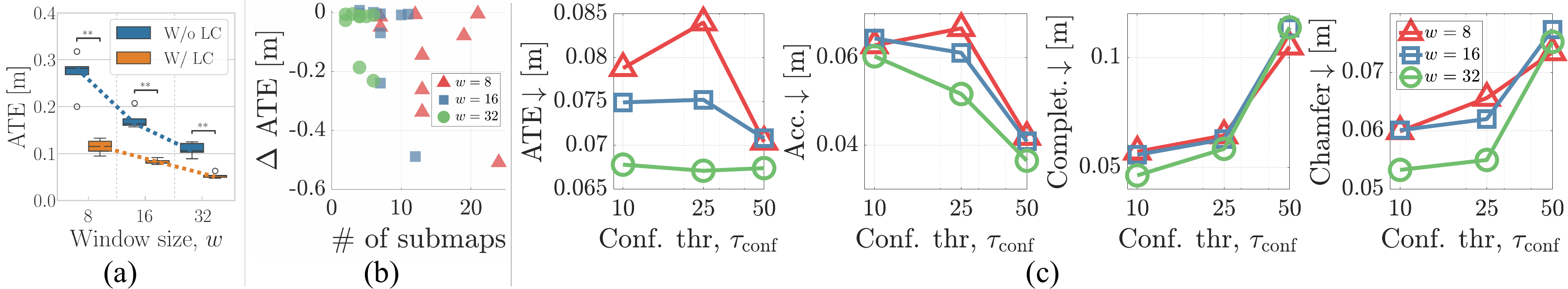}
    \end{subfigure}
    \caption{Ablation studies: (a) Effect of loop closure (LC) on absolute trajectory error (ATE) across different window sizes, $w$, in TUM~\cite{Sturm12iros-TUM-RGB-D}, 
      where ** annotations indicate measurements with $10^{-3} < $ $p$-value $\leq 10^{-2}$ after a paired $t$-test.
    (b)~Reduction in ATE achieved by incorporating loop closure in TUM, highlighting that as the number of submaps increases, our \SLfour-based optimization leads to greater reductions in ATE.
    (c)~Performance changes with respect to the confidence threshold $\tau_\mathrm{conf}$ in ATE, accuracy, completion, and Chamfer distance under varying window sizes in 7-Scenes~\cite{Shotton13cvpr}.}
    \label{fig:ablation}
\end{figure}

In \Cref{fig:ablation} we provide multiple ablation studies which show the following for three different 
submap sizes ($w=8,16,32$): (a) improved pose accuracy of \name when loop closures 
are leveraged along with showing a tight statistical spread of results from averaging 5 runs from our experiments, (b) that loop closures 
generally lead to increasing reduction in pose error as the number of submaps increases since there are an increased number 
of loop closures, (c) the effect of different values of 
$\tau_\mathrm{conf}$, which as expected, larger values lead to higher accuracy on dense reconstruction and lower completion, with our 
default value of 25\% showing an appropriate balance.

\section{Limitations}
\label{sec:limitations}

We have presented a new type of SLAM system that addresses the issue of projective ambiguity from an underlying feed-forward 
scene reconstruction method (in our case, VGGT). As creating a factor graph that optimizes on the \SLfour manifold is a 
new paradigm for the SLAM problem, it leaves much ground for further improvements. In particular, the estimation of 
the full 15-DOF homography matrix is degenerate in the case of planar points, which can lead to unstable solutions as we 
have observed in the planar floor scene of the TUM dataset. Our current implementation of homography 
using points from VGGT is also vulnerable to outliers. While we use a 5-point RANSAC to reduce 
this issue, the presence of a high outlier ratios or adversarial outliers (which are present due to local consistency 
of points in VGGT) can cause incorrect homography estimates as discussed in \Cref{sec:pose}. 
The ray-based matching in \mr-SLAM provides robustness to errors in depth measurements, 
and a similar method can potentially be adapted for homography estimation. 
Additionally, 15 DOF give rise to 
added opportunity of scene drift. While our addition of loop closures substantially corrects drift, an inaccurate relative homography 
estimate or long time between loop closures can cause not just drift in scale, rotation, and translation seen in classical SLAM, but also 
in scene perspective, which opens up an interesting area of research into further optimization into \SLfour-based SLAM. 

\section{Conclusion}

In this study, we have leveraged VGGT, a feed-forward reconstruction model, to incrementally construct a dense 
map from uncalibrated monocular cameras, proposing a novel SLAM framework called \textit{VGGT-SLAM}, which locally and 
globally (through loop closures) aligns submaps from VGGT. 
By exploring VGGT's geometric understanding though the lens of classical multi-view computer vision, we have shown that 
in the general case, these submaps must be aligned with a projective transformation, and in doing so we 
have created the first factor graph SLAM system optimized on the \SLfour manifold. 
In future work we will further investigate conditions under which \Simthree optimization suffices and 
investigate techniques to actively employ both \Simthree and \SLfour optimization in a unified system 
to enable a more robust SLAM system for 
real-time performance.

\section*{Acknowledgments}
This work is supported in part 
by the NSF Graduate Research Fellowship Program under Grant 2141064, the ONR RAPID program, and 
the National Research Foundation of Korea (NRF) grant funded by the Korea government (MSIT) (No. RS-2024-00461409). 
The authors would like to gratefully acknowledge Riku Murai for assisting us with benchmarking.

{
  
\small
\bibliographystyle{ieee_fullname}

}

\newpage

\appendix

\section{Tangent space of \SLfour}\label{appendix:tangent_space}
Here, we provide the explicit 15 generators, $\generator_k \; \forall k:\{1:15\}$, of \SLfour,
which allow us to relate the Lie algebra $\slfour$ to the Lie group $\SLfour$. 

The tangent space of \SLfour consists of all $4 \times 4$ real matrices with zero trace.
Thus, there are 15 generators, $\generator_k$, where 12 of them are defined as $\ME_{ab}$ for $a \neq b$ where 1 is in the $(a,b)$ entry and 0, elsewhere.
The remaining three generators are $\genDiag_1 = \mathrm{diag}(1, -1, 0, 0)$, $\genDiag_2 = \mathrm{diag}(0, 1, -1, 0)$, $\genDiag_3 = \mathrm{diag}(0, 0, 1, -1)$.
Explicitly, the generators are as follows:

{
\small
\[
\begin{aligned}
\generator_1 = \genElem_{01} &= 
\begin{pmatrix}
0 & 1 & 0 & 0 \\
0 & 0 & 0 & 0 \\
0 & 0 & 0 & 0 \\
0 & 0 & 0 & 0 
\end{pmatrix}, & 
\generator_2 = \genElem_{02} &= 
\begin{pmatrix}
0 & 0 & 1 & 0 \\
0 & 0 & 0 & 0 \\
0 & 0 & 0 & 0 \\
0 & 0 & 0 & 0 
\end{pmatrix}, &
\generator_3 = \genElem_{03} &= 
\begin{pmatrix}
0 & 0 & 0 & 1 \\
0 & 0 & 0 & 0 \\
0 & 0 & 0 & 0 \\
0 & 0 & 0 & 0 
\end{pmatrix}, \\[12pt]
\generator_4 = \genElem_{10} &= 
\begin{pmatrix}
0 & 0 & 0 & 0 \\
1 & 0 & 0 & 0 \\
0 & 0 & 0 & 0 \\
0 & 0 & 0 & 0 
\end{pmatrix}, &
\generator_5 = \genElem_{12} &= 
\begin{pmatrix}
0 & 0 & 0 & 0 \\
0 & 0 & 1 & 0 \\
0 & 0 & 0 & 0 \\
0 & 0 & 0 & 0 
\end{pmatrix}, &
\generator_6 = \genElem_{13} &= 
\begin{pmatrix}
0 & 0 & 0 & 0 \\
0 & 0 & 0 & 1 \\
0 & 0 & 0 & 0 \\
0 & 0 & 0 & 0 
\end{pmatrix}, \\[12pt]
\generator_7 = \genElem_{20} &= 
\begin{pmatrix}
0 & 0 & 0 & 0 \\
0 & 0 & 0 & 0 \\
1 & 0 & 0 & 0 \\
0 & 0 & 0 & 0 
\end{pmatrix}, &
\generator_8 = \genElem_{21} &= 
\begin{pmatrix}
0 & 0 & 0 & 0 \\
0 & 0 & 0 & 0 \\
0 & 1 & 0 & 0 \\
0 & 0 & 0 & 0 
\end{pmatrix}, &
\generator_9 = \genElem_{23} &= 
\begin{pmatrix}
0 & 0 & 0 & 0 \\
0 & 0 & 0 & 0 \\
0 & 0 & 0 & 1 \\
0 & 0 & 0 & 0 
\end{pmatrix}, \\[12pt]
\generator_{10} = \genElem_{30} &= 
\begin{pmatrix}
0 & 0 & 0 & 0 \\
0 & 0 & 0 & 0 \\
0 & 0 & 0 & 0 \\
1 & 0 & 0 & 0 
\end{pmatrix}, &
\generator_{11} = \genElem_{31} &= 
\begin{pmatrix}
0 & 0 & 0 & 0 \\
0 & 0 & 0 & 0 \\
0 & 0 & 0 & 0 \\
0 & 1 & 0 & 0 
\end{pmatrix}, &
\generator_{12} = \genElem_{32} &= 
\begin{pmatrix}
0 & 0 & 0 & 0 \\
0 & 0 & 0 & 0 \\
0 & 0 & 0 & 0 \\
0 & 0 & 1 & 0 
\end{pmatrix}, \\[12pt]
\generator_{13} = \genDiag_{1} & = 
\begin{pmatrix}
1 & 0 & 0 & 0 \\
0 & -1 & 0 & 0 \\
0 & 0 & 0 & 0 \\
0 & 0 & 0 & 0 
\end{pmatrix}, &
\generator_{14} = \genDiag_{2} & = 
\begin{pmatrix}
0 & 0 & 0 & 0 \\
0 & 1 & 0 & 0 \\
0 & 0 & -1 & 0 \\
0 & 0 & 0 & 0 
\end{pmatrix}, &
\generator_{15} = \genDiag_{3} & = 
\begin{pmatrix}
0 & 0 & 0 & 0 \\
0 & 0 & 0 & 0 \\
0 & 0 & 1 & 0 \\
0 & 0 & 0 & -1 
\end{pmatrix}.
\end{aligned}
\]
}

Thus, as briefly explained in \Cref{sec:backend},
the relation between the Lie algebra, $\vecstate^{\wedge} \in \slfour$, and the Lie group $\projTf \in \SLfour$ is given by:

\begin{equation}
  \projTf = \exp\left(\vecstate^{\wedge} \right) = \exp\left( \sum_{k=1}^{15} \vecstate_k \, \generator_k \right).
\end{equation}

\section{Extra Quantitative Results}
\label{sec:extra_quantitative}
We provide addition results of evaluating on the 7-Scenes~\cite{Shotton13cvpr} and TUM RGB-D~\cite{Sturm12iros-TUM-RGB-D} datasets 
where we experiment with different submap sizes (\Cref{sec:submap_sizes}) 
and show the number of submaps and loop closures per scene (\Cref{sec:num_submaps}). 

\subsection{Evaluation with different submap sizes}
\label{sec:submap_sizes}

Here we show results for the 7-Scenes and TUM RGB-D datasets in Tables~\ref{tab:7scenes_ate_all} and \ref{tab:tum_ate_all} with 
different submap sizes ($w=8,16,32$). For 7-Scenes, we also include results for $w=1$. Recall 
that $w$ is the size of new frames in the submap, so in the case of $w=1$, each submap has one new frame, one frame from the prior 
submap, and up to one extra frame from loop closures. For small submap size of $w=1$, the backend becomes numerically unstable 
for some TUM scenes (consistently floor and 360) preventing an estimated alignment, and thus we do not include the $w=1$ for TUM. 
This is due to reasons discussed in \Cref{sec:limitations}. Particularly, for the floor scene there are a large portion of images 
which only view a planar scene 
which makes the estimation of the full 15-DOF homography matrix degenerate, and for the 360 scene, using a small submap size such as 
$w=1$ is likely to encounter a pure rotation which can result in less accurate depth measurements from VGGT and hence reduced accuracy in 
our estimate of relative homography.

\begin{table}[h!]
  \vspace{-0.5em}
  \centering
  \caption{Root mean square error~(RMSE) of absolute trajectory error (ATE) on 7-Scenes~\cite{Shotton13cvpr} (unit: m).
  The~*~symbol indicates that the baseline is evaluated in the uncalibrated mode, all \name configurations are evaluated in uncalibrated mode.
  Green is best and light green is second best.}\label{tab:7scenes_ate_all}
  \scriptsize
  \begin{tabular}{l|lcccccccc} %
  \toprule
  \multirow{2}{*}{} & \multirow{2}{*}{Method} & \multicolumn{7}{c}{Sequence} &  \multirow{2}{*}{Avg}  \\ \cmidrule(lr){3-9}
                    & &\texttt{chess} &\texttt{fire} &\texttt{heads} &\texttt{office} &\texttt{pumpkin} &\texttt{kitchen} &\texttt{stairs} & \\
  \midrule
  \parbox[t]{1mm}{\multirow{8}{*}{\rotatebox[origin=c]{90}{Uncalib.}}} 
                    &DROID-SLAM*~\cite{Teed21nips-DROID-SLAM} & 0.047 & 0.038 & 0.034 & 0.136 & 0.166 & 0.080 &  \secondc 0.044 & 0.078  \\
                    &\mr-SLAM*~\cite{Murai24arxiv-Mast3RSLAM} & 0.063 & 0.046 & 0.029 & \firstc 0.103 & \firstc {0.114} &0.074 & \firstc 0.032 & \firstc 0.066 \\
                    &Ours ($\Simthree, w=1$) &  0.047 & \firstc 0.025 & 0.032 &  0.113 & 0.138 & \firstc 0.050 & 0.083 & 0.070  \\
                    &Ours ($\Simthree, w=8$)  & 0.039 & 0.027 & 0.020 & 0.108 & 0.144 &  0.053 & 0.080 & \secondc 0.067 \\
                    &Ours ($\Simthree, w=16$) & 0.037 & 0.027 & 0.021 & 0.107 & 0.135 &  0.051 & 0.093 & \secondc 0.067 \\
                    &Ours ($\Simthree, w=32$) & 0.037 & \secondc 0.026 & \firstc 0.018 &  0.104 & \secondc 0.133 & 0.061 & 0.093 & \secondc 0.067 \\
                    &Ours ($\SLfour, w=1$) &  0.089 & 0.046 & 0.072 &  0.119 & 0.147 & 0.055 & 0.100 & 0.090  \\
                    &Ours ($\SLfour, w=8$) &  0.041 & 0.060 & 0.043 & 0.106 & 0.206 & 0.054 &  0.078 & 0.084  \\
                    &Ours ($\SLfour, w=16$) & \firstc 0.036 & 0.065 & 0.037 & 0.107 & 0.139 & \firstc 0.050 & 0.093 & 0.075  \\
                    &Ours ($\SLfour, w=32$) & \firstc 0.036 & 0.028 & \firstc 0.018 & \firstc 0.103 & \secondc 0.133 & 0.058 & 0.093 & \secondc 0.067  \\
  \bottomrule
  \end{tabular}
\end{table}

\begin{table}[h!]
  \vspace{-0.5em}
  \centering
  \caption{Root mean square error~(RMSE) of absolute trajectory error (ATE) on TUM RGB-D~\cite{Sturm12iros-TUM-RGB-D} (unit: m).
  The~*~symbol indicates that the baseline is evaluated in the uncalibrated mode, all \name configurations are evaluated in uncalibrated mode.
  Green is best and light green is second best.}\label{tab:tum_ate_all}
  \scriptsize
  \begin{tabular}{l|lcccccccccc} 
  \toprule
  \multirow{2}{*}{} & \multirow{2}{*}{Method} & \multicolumn{9}{c}{Sequence} &  \multirow{2}{*}{Avg}  \\ \cmidrule(lr){3-11}
  & &\texttt{360} &\texttt{desk} &\texttt{desk2} &\texttt{floor} &\texttt{plant} &\texttt{room } &\texttt{rpy} &\texttt{teddy} &\texttt{xyz} & \\
  \midrule
  \parbox[t]{1mm}{\multirow{8}{*}{\rotatebox[origin=c]{90}{Uncalib.}}} &DROID-SLAM*~\cite{Teed21nips-DROID-SLAM} &0.202 & 0.032 &0.091 & 0.064 &0.045 &0.918 &0.056 &0.045 & \firstc 0.012 &0.158 \\
  &\mr-SLAM*~\cite{Murai24arxiv-Mast3RSLAM} &\firstc {0.070} & 0.035 & 0.055 & \secondc 0.056 &0.035 & 0.118 & 0.041 &0.114 & 0.020 & 0.060 \\
  &Ours~(\Simthree, $w=8$)  &\firstc 0.070 & \secondc 0.026 & \firstc 0.030 & \firstc 0.048 & 0.026 & \secondc 0.081 & \firstc 0.024 & 0.035 & 0.015 & \firstc 0.040 \\
  &Ours~(\Simthree, $w=16$) & 0.112 & 0.045 & 0.123 & 0.261 & \firstc 0.022 & 0.137 & 0.044 & 0.044 & 0.016 & 0.089 \\
  &Ours~(\Simthree, $w=32$) & 0.123 & 0.040 & 0.055 & 0.254 & \firstc 0.022 & \firstc 0.088 & 0.041 & \firstc 0.032 & 0.016 & 0.074 \\
  &Ours~(\SLfour, $w=8$)  & 0.179 & 0.046 & 0.095 & 0.210 & 0.033 & 0.272 & 0.038 & 0.130 & 0.031  & 0.115 \\
  &Ours~(\SLfour, $w=16$) & 0.147 & 0.032 & 0.087 & 0.158 & 0.027 & 0.150 & 0.037 & 0.069 & 0.035 & 0.083 \\
  &Ours~(\SLfour, $w=32$) & 0.071 & \firstc 0.025 & \secondc 0.040 & 0.141 & 0.023 & 0.102 & \secondc 0.030 & \secondc 0.034 & \secondc 0.014 & \secondc 0.053 \\
  \bottomrule
  \end{tabular}
  \vspace{-1.0em}
\end{table}

\begin{table}[h!]
  \vspace{-1.2em}
  \centering
  \caption{Dense reconstruction evaluation on 7-Scenes~\cite{Shotton13cvpr} (unit: m).}
  \setlength{\tabcolsep}{2.0pt}
  \scriptsize
  \begin{tabular}{l|lcccc}
    \toprule
    \multirow{2}{*}{} & \multirow{2}{*}{Method} & \multicolumn{4}{c}{7-Scenes} \\ \cmidrule(lr){3-6}
     & & ATE\,$\downarrow$ & Acc.\,$\downarrow$ & Complet.\,$\downarrow$ & Chamfer\,$\downarrow$ \\
    \midrule
    \parbox[t]{2mm}{\multirow{7}{*}{\rotatebox[origin=c]{90}{Uncalib.}}}
    & \mr-SLAM*~\cite{Murai24arxiv-Mast3RSLAM}  & \firstc 0.066 & 0.068 & \firstc 0.045 & 0.056  \\
    & Ours (\Simthree, $w=1$)  & 0.070 & 0.066 & \secondc 0.051 & 0.059 \\
    & Ours (\Simthree, $w=8$)  & \secondc 0.067 & 0.054 & 0.056 & \firstc 0.055 \\
    & Ours (\Simthree, $w=16$) & \secondc 0.067 & 0.054 &  0.058 & 0.056 \\
    & Ours (\Simthree, $w=32$) & \secondc 0.067 & \firstc 0.052 & 0.062 & 0.057 \\
    & Ours (\SLfour, $w=1$)  & 0.090 & 0.080 & 0.068 & 0.074 \\
    & Ours (\SLfour, $w=8$)  & 0.084 & 0.067 & 0.065 & 0.066 \\
    & Ours (\SLfour, $w=16$) & 0.075 & 0.061 & 0.063 & 0.060 \\
    & Ours (\SLfour, $w=32$) & \secondc 0.067 & \firstc 0.052 & 0.058 & \firstc 0.055 \\
    \bottomrule
  \end{tabular}
  \vspace{-1em}
  \label{tab:7scenes_recon_all}
\end{table}

\subsection{Number of submaps per scene}
\label{sec:num_submaps}

As a reference, in Tables~\ref{tab:7scenes_window_size} and \ref{tab:tum_window_size} we show the number of total submaps in 
each scene for 7-Scenes and TUM RGB-D for different values 
of experimented submap size, $w$, and also show the number of loop closures in each scene.

\begin{table}[!hbt]
  \centering
  \caption{Window size $w$ and corresponding submap and loop closure counts when $\thrnumloop = 1$, shown as ``\#~of~submaps~(\#~of~loops)''.
  }\label{tab:7scenes_window_size}
  \scriptsize
  \begin{tabular}{c|ccccccc} 
  \toprule
  \multirow{2}{*}{Window size, $w$} & \multicolumn{7}{c}{Sequences in 7-Scenes~\cite{Sturm12iros-TUM-RGB-D}} \\ \cmidrule(lr){2-8}
                               & \texttt{chess} &\texttt{fire} &\texttt{heads} &\texttt{office} &\texttt{pumpkin} &\texttt{kitchen} &\texttt{stairs} \\ \midrule
  1  &  29 (11) & 50 (46) & 62 (49) & 58 (55) & 43 (37) & 43 (38) & 14 (12)   \\
  8  &  4 (0) & 7 (4) & 8 (3) & 8 (4) & 6 (0) & 6 (2) & 2 (0)   \\
  16 &  2 (0) & 4 (1) & 4 (2) & 4 (2) & 3 (0) & 3 (1) & 1 (0)  \\
  32 &  1 (0) & 2 (0) & 2 (0) & 2 (0) & 2 (0) & 2 (0) & 1 (0)  \\
  \bottomrule
  \end{tabular}
\end{table}

\begin{table}[!hbt]
    \centering
    \caption{Window size $w$ and corresponding submap and loop closure counts when $\thrnumloop = 1$, shown as ``\#~of~submaps~(\#~of~loops)''.
    }\label{tab:tum_window_size}
    \scriptsize
    \begin{tabular}{c|ccccccccc} 
    \toprule
    \multirow{2}{*}{Window size, $w$} & \multicolumn{9}{c}{Sequences in TUM-RGB-D~\cite{Sturm12iros-TUM-RGB-D}} \\ \cmidrule(lr){2-10}
    &\texttt{360} &\texttt{desk} &\texttt{desk2} &\texttt{floor} &\texttt{plant} &\texttt{room } &\texttt{rpy} &\texttt{teddy} &\texttt{xyz} \\ \midrule
    1  &  168 (151) & 54 (42) & 98 (84) & 99 (87) & 102 (92) & 186 (162) & 95 (89) & 146 (125) & 56 (54) \\
    8  &  21 (4) & 7 (4) & 13 (7) & 13 (3) & 13 (5) & 24 (7) & 12 (10) & 19 (9) & 7 (5) \\
    16 &  11 (2) & 4 (2) &  7 (4) &  7 (2) &  7 (2) & 12 (4) & 6 (4) & 10 (4) & 4 (2)  \\
    32 &  6  (1) & 2 (0) &  4 (2) &  4 (1) &  4 (2) &  6 (2) & 3 (1) &  5 (2) & 2 (0) \\
    \bottomrule
    \end{tabular}
  \end{table}

\newpage

\section{Extra Qualitative Results}

\subsection{Extra examples of \SLfour versus \Simthree}
While we have mentioned that the \Simthree version of \name often provides high quality reconstructions, 
here we provide additional examples of cases where \Simthree is not sufficient and \SLfour is necessary to achieve 
consistent alignment across submaps. 

\begin{figure}[!hbt]
  \centering
  \includegraphics[width=\linewidth]{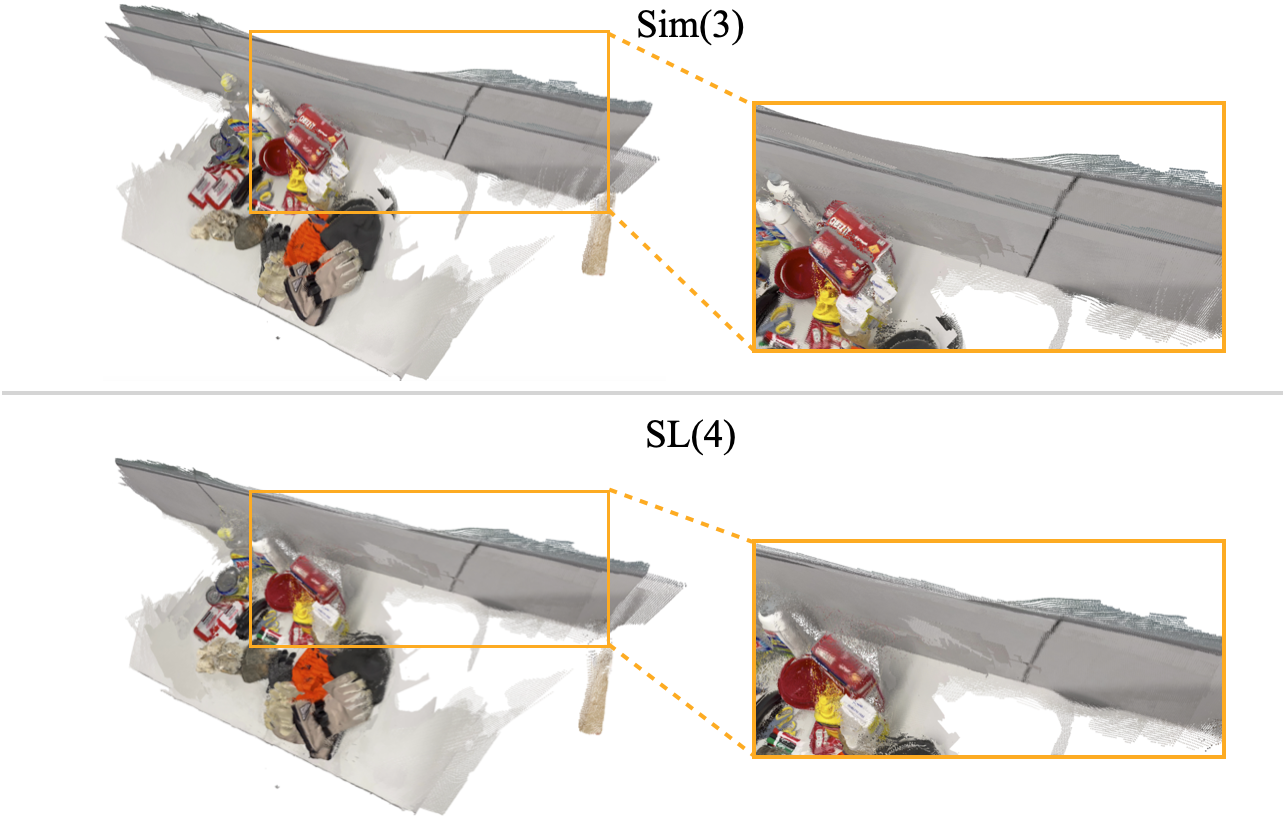}
  \caption{Example on a tabletop scene showing \Simthree is unable to align the submaps while \SLfour is able to correct 
  for projective ambiguity. Here $w=32$ and $\thrdisparity = 50$.}
  \label{fig:sim3_1}
\end{figure}

\begin{figure}[!hbt]
  \centering
  \includegraphics[width=\linewidth]{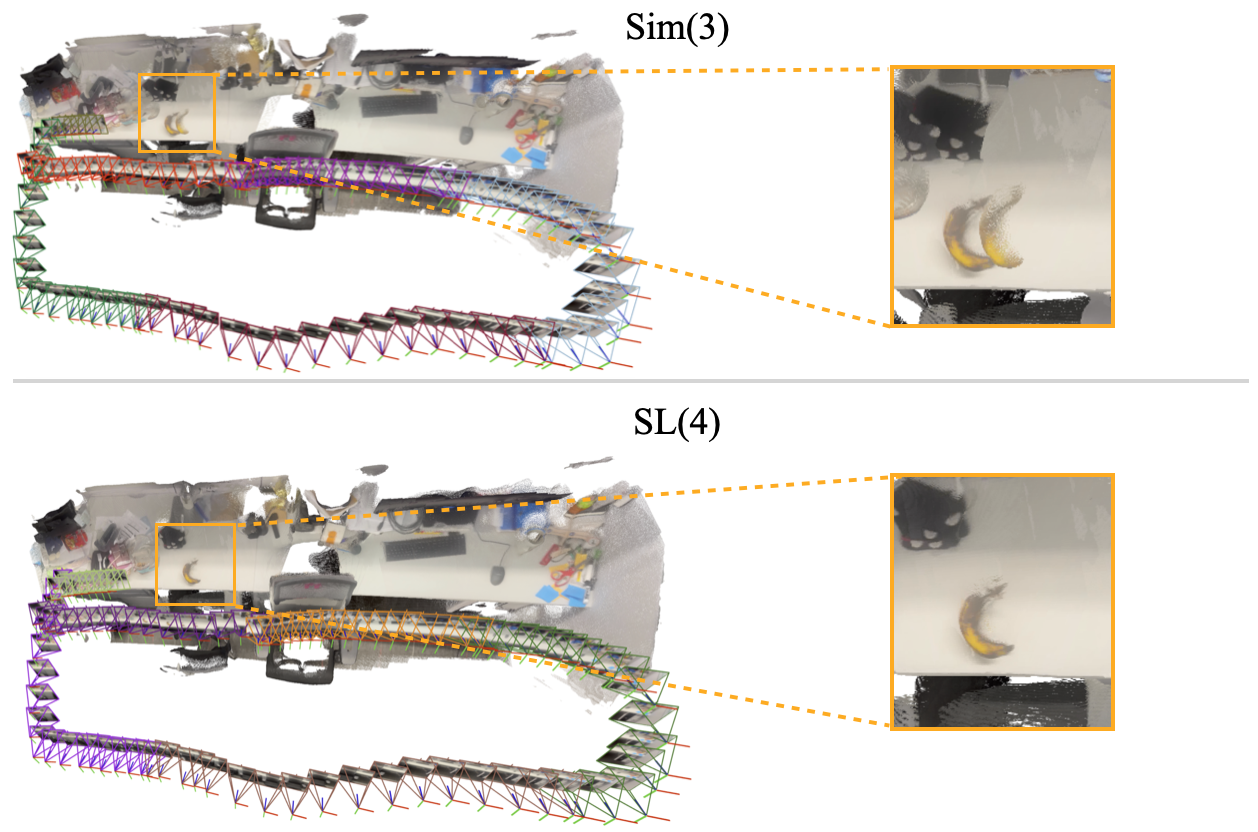}
  \caption{Example on a tabletop scene showing \Simthree is unable to align the submaps while \SLfour is able to correct 
  for projective ambiguity. The true scene only has one banana, but the \Simthree reconstruction shows 
  a hallucination of two caused by misalignment. Camera pose estimates are colored by submap. Here $w=16$ and $\thrdisparity = 50$.}
  \label{fig:sim3_2}
\end{figure}

\begin{figure}[!hbt]
  \centering
  \includegraphics[width=\linewidth]{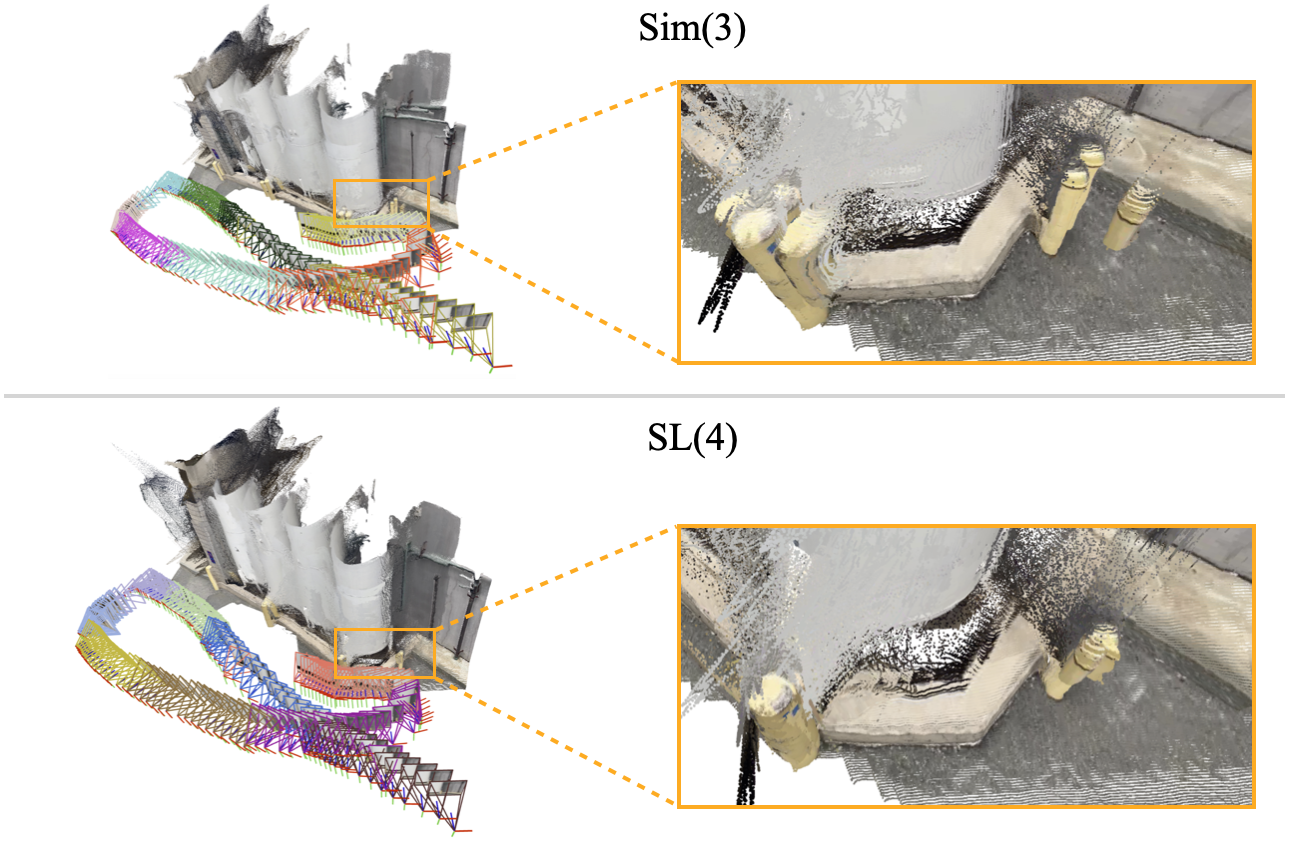}
  \caption{Example on an outdoor scene with yellow bollards surrounding tanks showing \Simthree is unable to 
  align the submaps while \SLfour is able to correct 
  for projective ambiguity. The true scene has single bollards spaced around the tanks while the \Simthree scene 
  hallucinates clusters of bollards due to misalignment. Here $w=16$ and $\thrdisparity = 25$.}
  \label{fig:sim3_3}
\end{figure}

\subsection{7-Scenes Qualitative Results}

Here we provide additional visualizations of scene reconstructions from the 7-Scenes dataset experiments 
for \name with \SLfour. 
We use the default parameters from \Cref{sec:experiments}. 

\begin{figure}[!hbt]
  \centering
  \includegraphics[width=\linewidth]{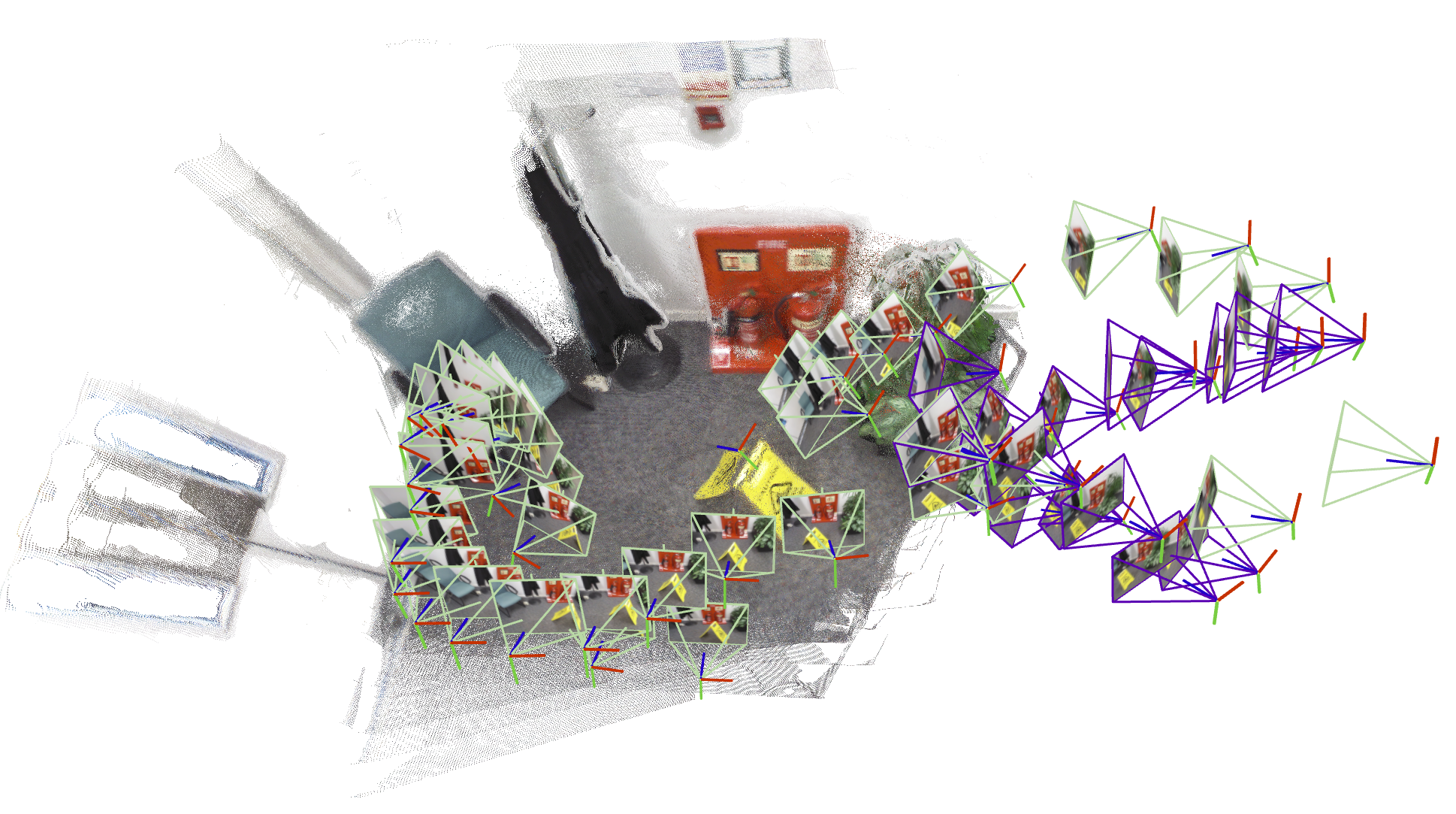}
  \caption{Visualization of reconstruction on 7-Scenes \texttt{fire} scene with 2 submaps. Camera pose estimates are colored by submap.}
  \label{fig:7scenes_fire}
\end{figure}

\begin{figure}[!hbt]
  \centering
  \includegraphics[width=\linewidth]{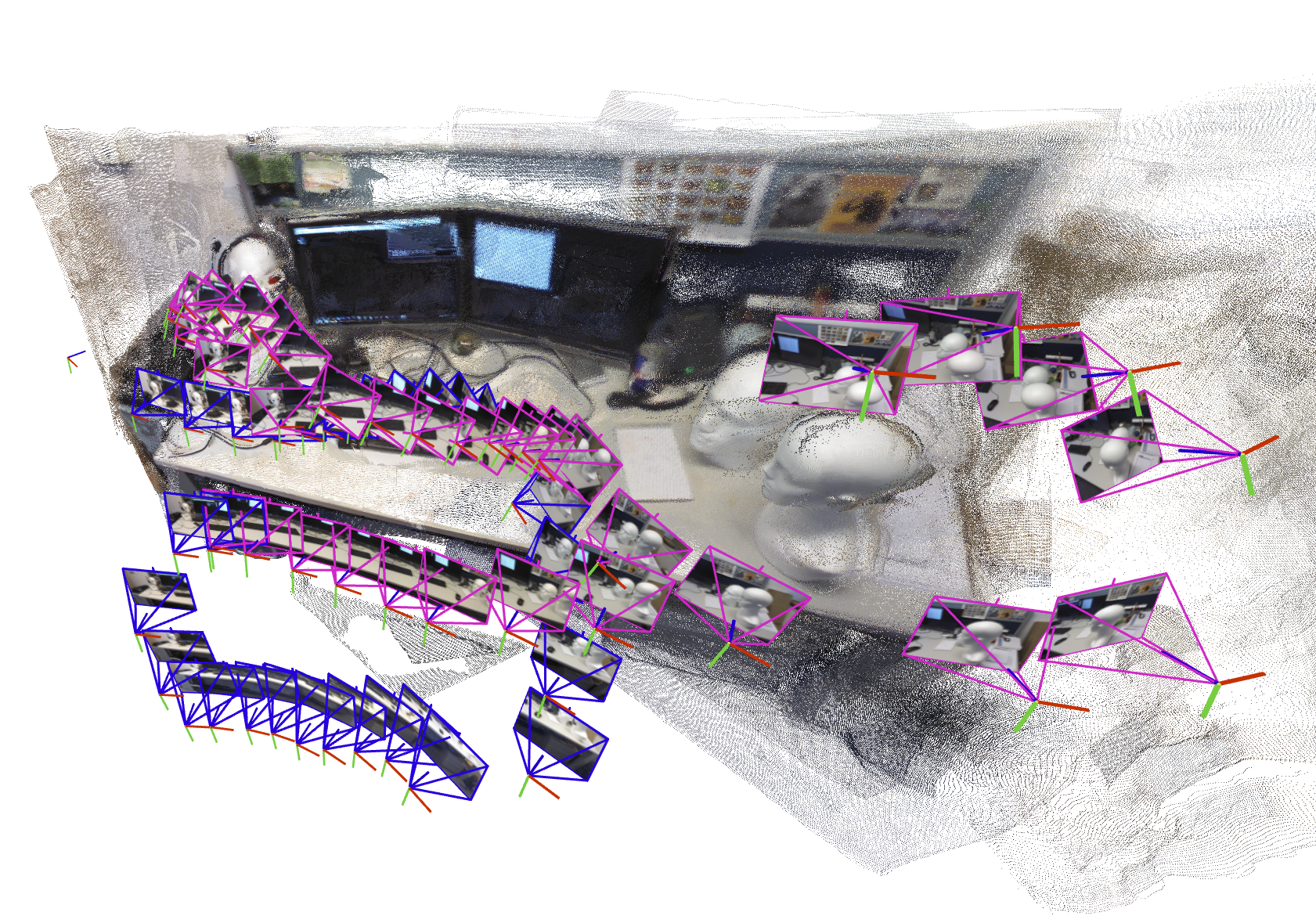}
  \caption{Visualization of reconstruction on 7-Scenes \texttt{heads} scene with 2 submaps. Camera pose estimates are colored by submap. 
  Part of the scene in cropped for 
  visual clarity.}
  \label{fig:7scenes_heads}
\end{figure}

\subsection{TUM RGB-D Qualitative Results}

Here we provide additional visualizations of scene reconstructions from the TUM RGB-D dataset experiments 
for \name with \SLfour. 
We use the default parameters from \Cref{sec:experiments}. 

\begin{figure}[!hbt]
  \centering
  \includegraphics[width=\linewidth]{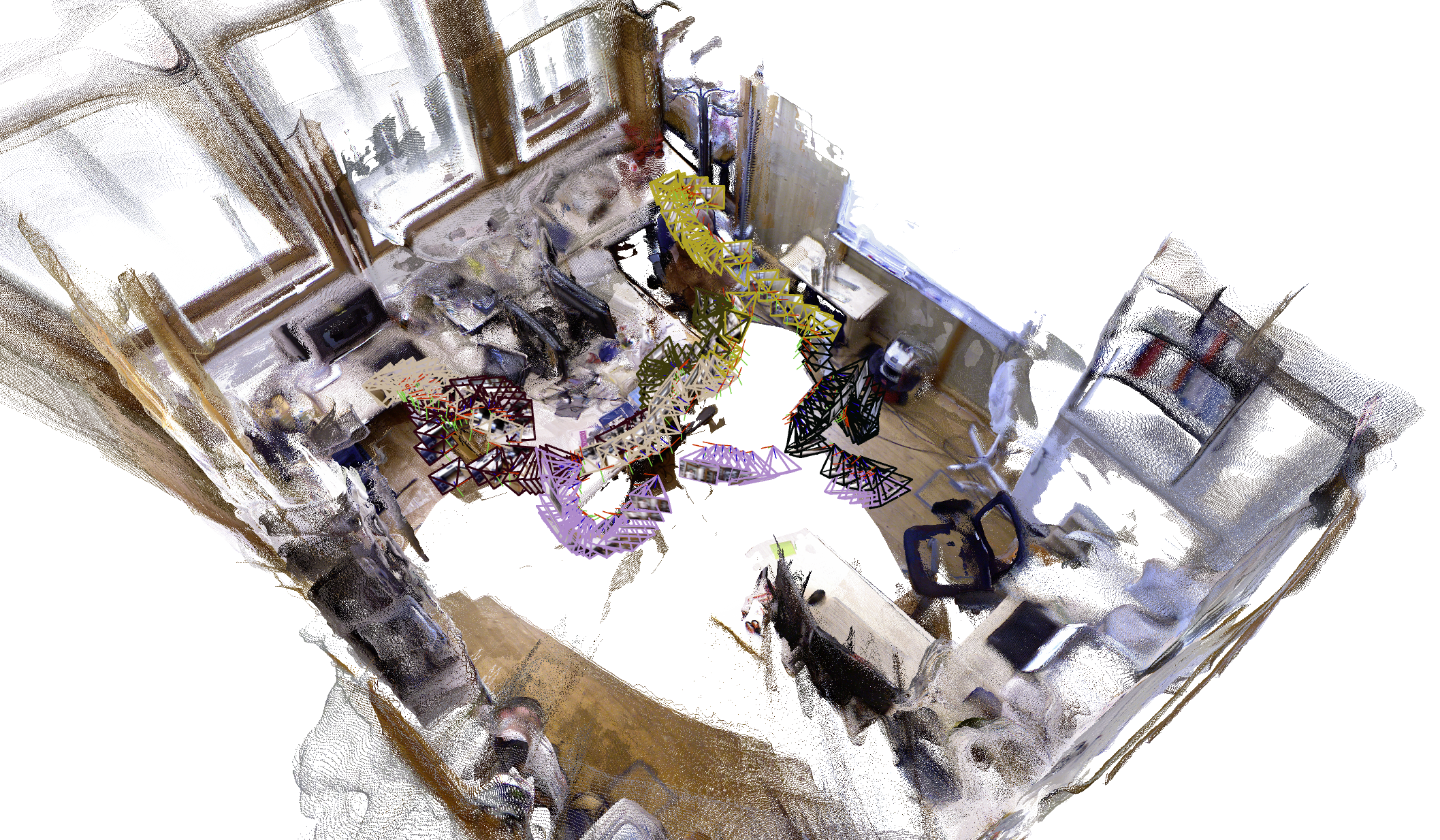}
  \caption{Visualization of reconstruction on TUM \texttt{room} scene with 6 submaps. Camera pose estimates are colored by submap. 
  Part of the scene in cropped for 
  visual clarity.}
  \label{fig:tum_room}
\end{figure}

\begin{figure}[!hbt]
  \centering
  \includegraphics[width=\linewidth]{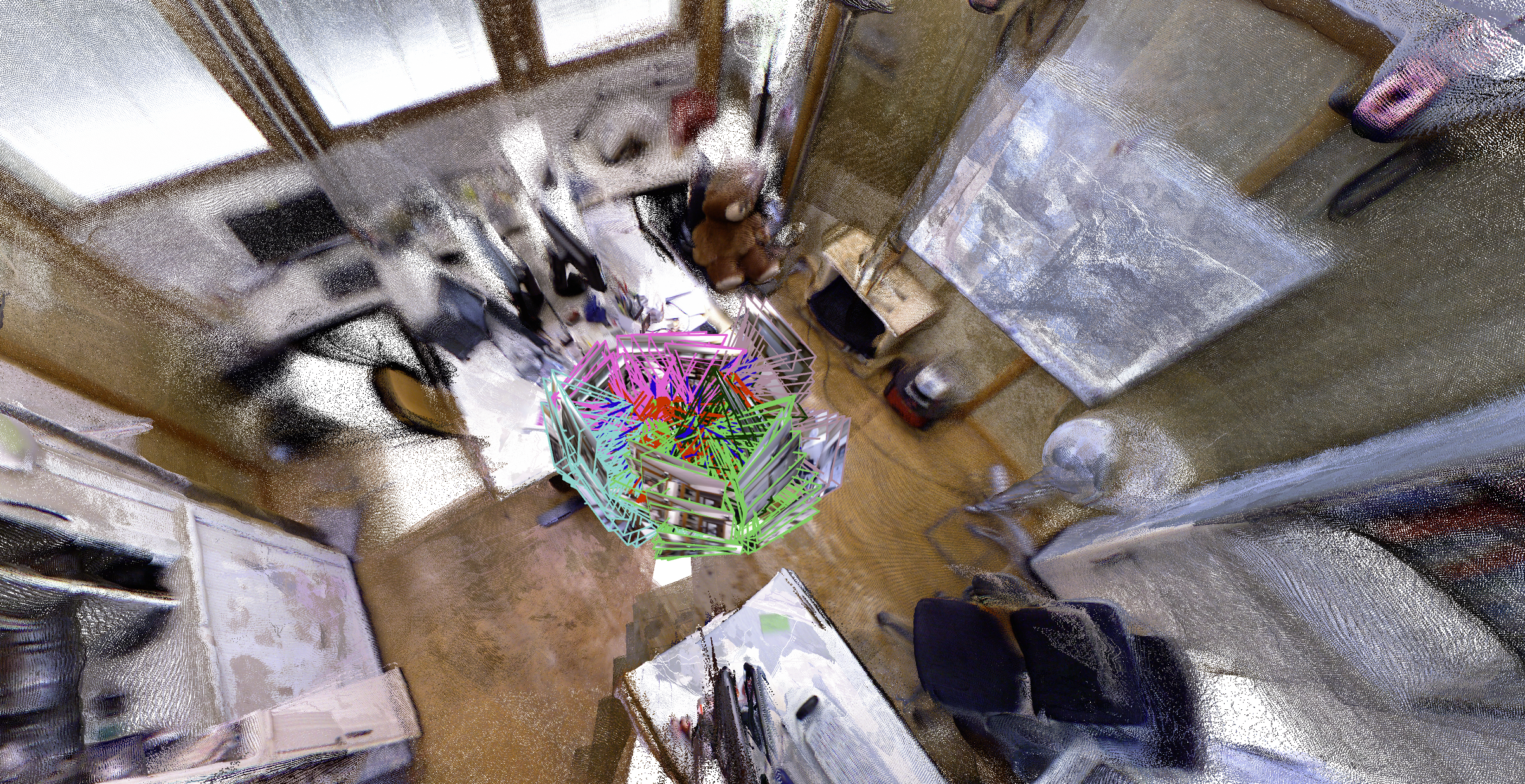}
  \caption{Visualization of reconstruction on TUM \texttt{360} scene with 6 submaps. Camera pose estimates are colored by submap. 
  Part of the scene in cropped for 
  visual clarity.}
  \label{fig:tum_360}
\end{figure}

\begin{figure}[!hbt]
  \centering
  \includegraphics[width=\linewidth]{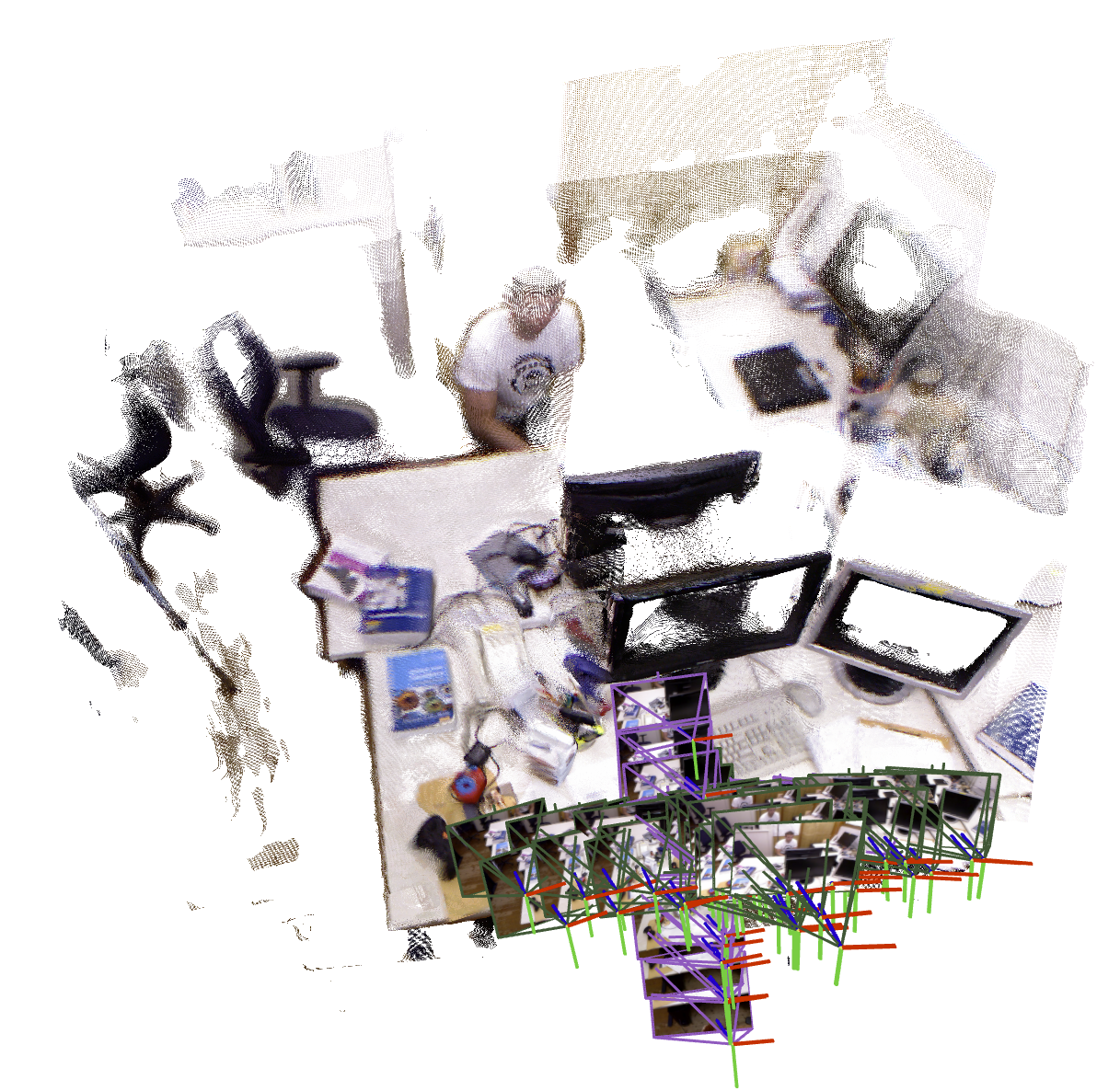}
  \caption{Visualization of reconstruction on TUM \texttt{xyz} scene with 2 submaps. Camera pose estimates are colored by submap.}
  \label{fig:tum_xyz}
\end{figure} %

\end{document}